\documentclass[master,english,final]{kaist-ucs}

\usepackage{enumitem}
\usepackage{listings}

\title[korean] {깊은 시각화를 이용한 파괴적 망각 극복}
\title[english]{Overcoming Catastrophic Forgetting by \\
Deep Visualization}

\author[korean] {뉴엔}{ 반 지앙}
\author[korean2] {뉴엔}{ 반 지앙}    
\author[chinese]{阮}{文 江}
\author[english]{Nguyen}{Van Giang}

\advisor[major]{김 대 영}{Daeyoung Kim}{signed}
\advisor[major2]{김 대 영}{Daeyoung Kim}{signed}    
\advisorinfo{Professor of School of Computing} 
\department{CS}{engineering}{a}

\studentid{20184658}

\referee[1]{김 대 영}
\referee[2]{오 혜 연}
\referee[3]{박 노 성}

\approvaldate{2020}{6}{11}

\refereedate{2020}{6}{11}

\gradyear{2020}

\begin{document}


   \thesisinfo

    \begin{abstract}
    Explaining the behaviors of deep neural networks, usually considered as black boxes, is critical especially when they are now being adopted over diverse aspects of human life. Taking the advantages of interpretable machine learning (interpretable ML), this work proposes a novel tool called Catastrophic Forgetting Dissector (or CFD) to explain catastrophic forgetting in continual learning settings. We also introduce a new method called Critical Freezing based on the observations of our tool. Experiments on ResNet articulate how catastrophic forgetting happens, particularly showing which components of this famous network are forgetting. Our new continual learning algorithm defeats various recent techniques by a significant margin, proving the capability of the investigation. Critical freezing not only attacks catastrophic forgetting but also exposes explainability.
    \end{abstract} 
     
    \begin{Engkeyword}
    Image captioning, Continual learning, Catastrophic forgetting, Interpretable ML
    \end{Engkeyword}

    \addtocounter{pagemarker}{1}                 
    \newpage

    \tableofcontents

    \listoftables

    \listoffigures



\chapter{Introduction}
\noindent
Regarding human evolution, life-long learning has been considered as one of the most crucial abilities, helping us develop more complicated skills throughout the lifetime. The idea of this learning strategy is hence deployed extensively by the deep learning community. Life-long learning (or continual learning) enables machine learning models to perceive new knowledge while simultaneously exposing backward-forward transfer, non-forgetting, or few-shot learning \cite{ling2019unified}. While the aforementioned properties are the ultimate goals for life-long learning systems, catastrophic forgetting or semantic drift naturally occurs in deep neural networks in life-long learning settings because they are vastly optimized upon gradient descent algorithm \cite{goodfellow2013empirical}. 

\begin{figure*}
\begin{center}
\includegraphics[scale=0.3]{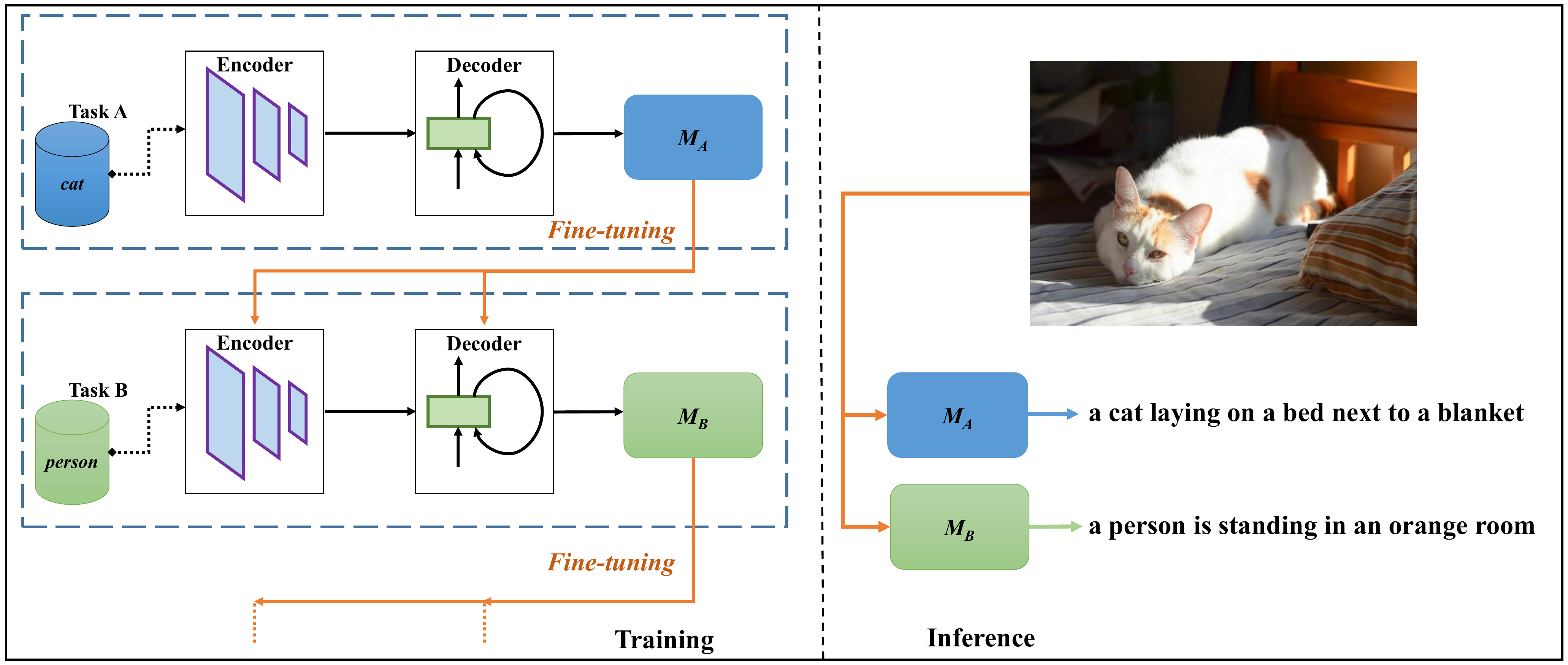}
\end{center}
   \caption{Catastrophic forgetting in continual learning setting.}
\label{fig:problem}
\end{figure*}

Catastrophic forgetting is defined as when we use a trained model on a given domain to address a new task. Due to adapting to the new data samples, the model forgets what it learned before on the old domain. As shown in Fig. \ref{fig:problem}, if we use fine-tuning for continual learning, the knowledge acquired from the previous task will be eradicated and the inference fails miserably. In fact, the description of the cat photo is shifted from ``\textit{a cat laying on a bed next to a blanket}'' to ``\textit{a person is standing in an orange room}''.

Although catastrophic forgetting is tough and undesirable, research on understanding this problem is rare amongst the deep learning community. The interest in understanding or measuring catastrophic forgetting does not commensurate with the number of research to deal with this problem. Kemker et al. \cite{kemker2018measuring} develop new metrics to help compare continual learning techniques fairly and directly. Nguyen et al. \cite{nguyen2019toward} study which properties cause the hardness for the learning process. By modeling the chosen properties using task space, they can estimate how much a model forgets in a sequential learning scenario, shedding light on factors affecting the error rate on a task sequence. They can not show us what is being forgotten or which components are forgetting inside the model, but revealing what properties of tasks trigger catastrophic forgetting. By comparison, our work focuses on understanding which components of a network are volatile corresponding to a given sequence of tasks to articulate catastrophic forgetting.

This research, in particular, introduces a novel approach to elaborate catastrophic forgetting by visualizing hidden layers in class-incremental learning (considered as the hardest scenario of continual learning). In this learning paradigm, the use of previous data is prohibited and instances of the incoming tasks are unseen. We develop a tool named Catastrophic Forgetting Dissector (or \texttt{CFD}) which automates the dissection of catastrophic forgetting, exactly pointing out which components, in a model, are causing the forgetting. We formally adopt Intersection over Union (IoU), a popular evaluation metric in detection and segmentation tasks which essentially computes the overlapping ratio between two frames, to measure the forgetting degree of deep neural networks in this work. The degree of forgetting is objectively measured after each class is added, thus giving us an intuition of how forgetting happens on a given part of the network.

To satisfy continual learning settings in image captioning we created a dataset named Split MS-COCO (a derivative of MS-COCO 2014) in which each task will come with a new class. For example, with an old task A of classes {cat, dog, and table}; the new tasks B will contain samples of class {person} only. When training task B, knowledge from task A is leveraged to help the current model generalize well on all 4 classes: {cat, dog, table, and person}. We define terminology ``clear image" to represent an image containing only one class out of 80 classes defined by MS-COCO in its reference captions to manipulate incremental steps correctly. Split MS-COCO enables us to comprehensively measure the forgetting on deep neural networks. This dataset is particularized in Chapter. \ref{sec:exp}.

\begin{figure*}[h!]
	\centering
	\includegraphics[scale=0.68]{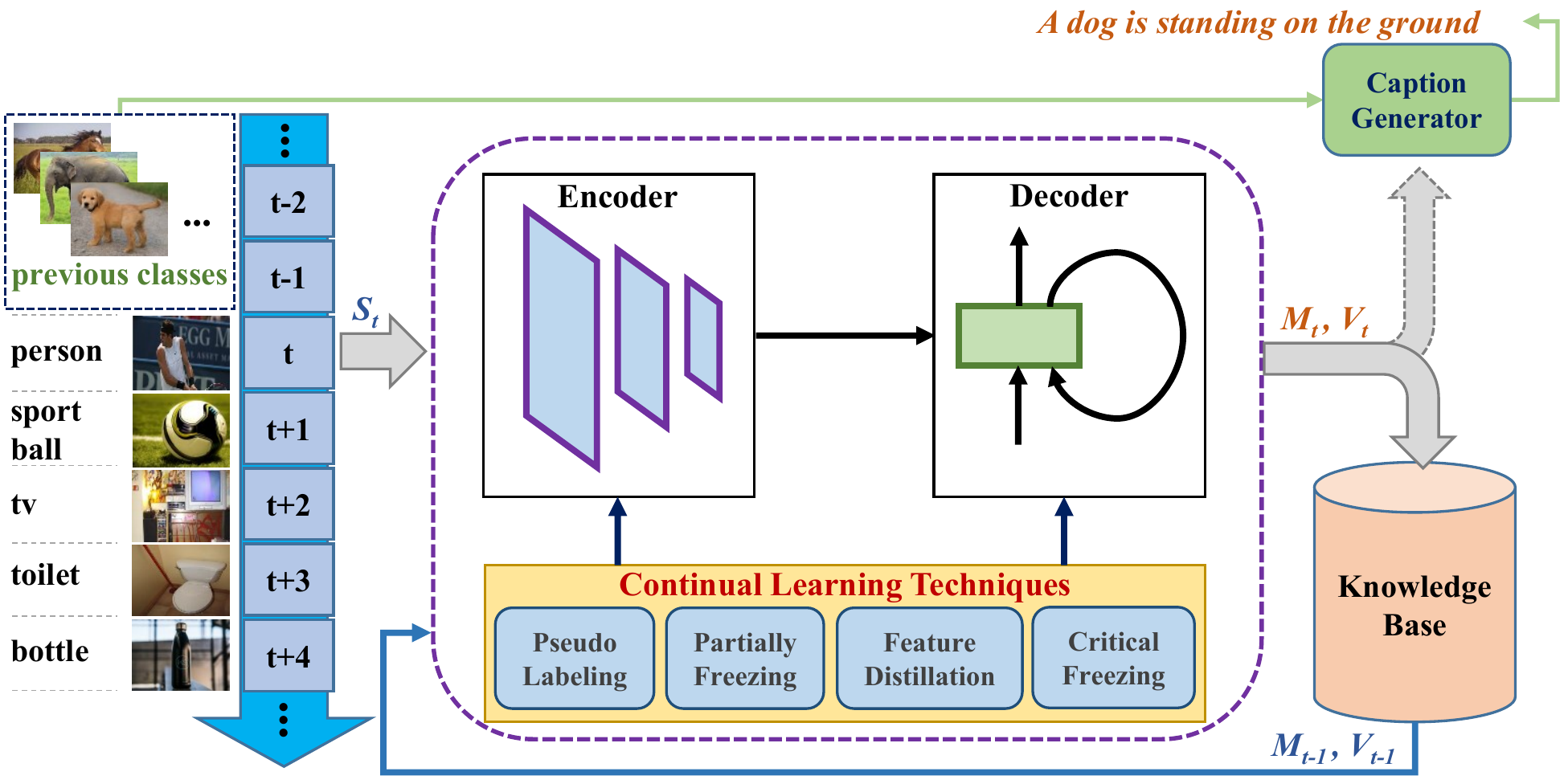}
	\caption{A scalable framework for continual image captioning.}
	\label{fig:framework}
\end{figure*}

We also propose \textbf{a novel framework} (as in Fig. \ref{fig:framework}) that is a scalable framework to combine encoder-decoder image captioning architecture with continual learning. The scalability of the framework is maintained as further continual learning techniques and more sophisticated encoder-decoder architectures can be smoothly integrated to increase the performance. Fine-tuning is considered as the baseline for comparison and evaluating the capability of strategies. 

From the results of the dissection, we try to infer the plastic components in the network to preserve the accumulated knowledge. Our new algorithm, critical freezing, protects these components by simply keeping weights unchanged while training the new network. The experiments demonstrate catastrophic forgetting in image captioning and the superiority of the proposed techniques over fine-tuning. The previously learned tasks are well performed, whilst new information is also well absorbed. Traditional metrics in image captioning, such as BLEU, ROUGE-L, CIDEr, and SPICE are calculated for quantitative assessments. Finally, we provide future directions and discussion over experiments to elaborate on the results qualitatively and quantitatively.

\noindent
\textbf{Contributions:} Our main contributions are as follows:
\begin{itemize}[noitemsep,nolistsep]
    \item Our work is the first attempt to address catastrophic forgetting in image captioning without the need for accessing data from existing tasks.
    \item We propose a scalable framework (including Split MS-COCO, baselines, experimental scenarios), which reconciles image captioning and continual learning in the class-incremental scenario.
    \item Our method is novel and pioneering in helping explain catastrophic forgetting in continual learning.
    \item We introduce a new approach to mitigate catastrophic forgetting based on the findings.
    \item Our extensive experiments demonstrate the efficacy of critical freezing and suggest recondite understanding about catastrophic forgetting. 
\end{itemize}

\newpage

\chapter{Background and Related Work}
\label{sec:relatedwork}
\section{Image Captioning}
Given an image, our goal is to generate the most suitable caption describing the content of an image. This task is rising as one of the most attractive domains in computer vision because of its exceptionally valuable applications such as aiding to the visually impaired, social media, digital assistant, or photo indexing. Image captioning models often follow the encoder-decoder architecture in which a powerful feature extractor (convolutional neural network - CNN) incorporates a sequence generator (recurrent neural network - RNN). Recently, attention mechanism has achieved huge successes in image captioning. From the introduction of the Neural Image Caption Generator (NIC) \cite{vinyals2015show}, various compelling techniques have been proposed to enhance image captioning \cite{lu2018neural, cornia2019show, aneja2018convolutional}. Experiments are frequently done on three benchmark datasets: Flickr8 \cite{hodosh2013framing}, Flickr30 \cite{plummer2015flickr30k}, and MS-COCO \cite{lin2014microsoft}. 

\begin{figure*}[h!]
	\centering
	\includegraphics[scale=1.0]{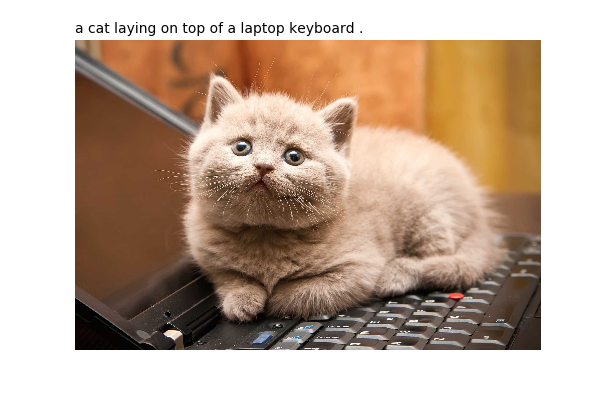}
	\caption{A synthetic caption is generated by deep learning.}
	\label{fig:cat}
\end{figure*}

The encoder compresses the input image to small images that represent features of target objects. For years, deep convolutional networks have captured features very well; as a result, transfer learning is leveraged in image captioning to produce high-quality immediate features. To do this, last layers of image classifiers \cite{he2016deep, szegedy2015going} are stripped away to expose immediate features that are then fed into a long-short term memory network (LSTM).
When generating a sequence, at each time step, the last generated word is taken into account to produce the next word, repeating until a special token (End of Sequence - EOS) appears. There is also a mapping process from the number generated by the network and a pre-built vocabulary. Fig. \ref{fig:cat} illustrates the prediction on a cat photo.

\section{Interpretable Machine Learning}
\begin{figure*}[h!]
	\centering
	\includegraphics[scale=0.3]{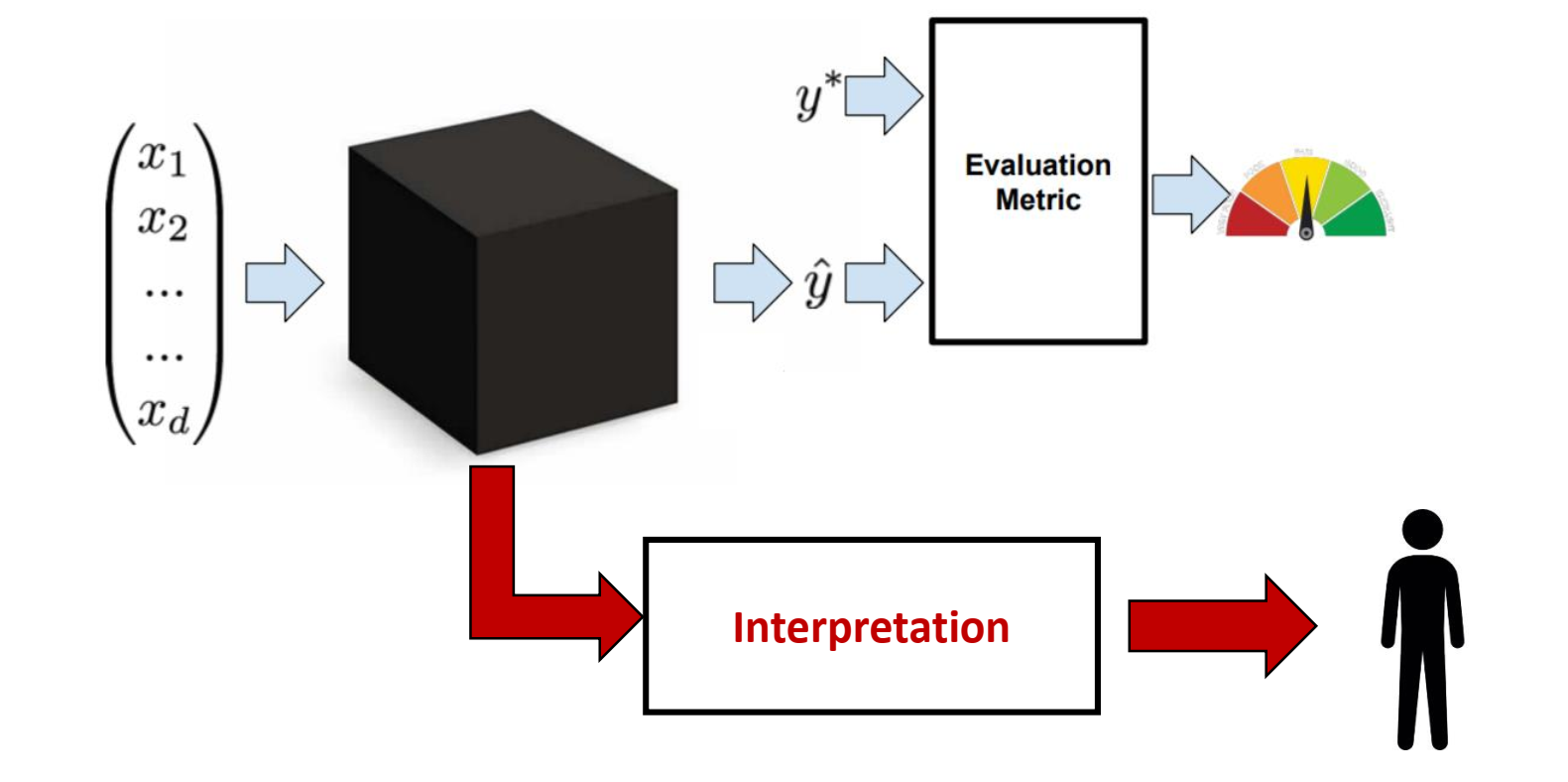}
	\caption{Interpretable ML helps open the black-box of deep learning.}
	\label{fig:black-box}
\end{figure*}

Interpretability is the degree to which human can understand the cause of a decision. Another one is: Interpretability is the degree to which a person can consistently predict the model’s result. The higher the interpretability of a machine learning model, the easier it is for someone to comprehend why certain decisions or predictions have been made. A model is better interpretable than another model if its decisions are easier for a human to comprehend than decisions from the other model. We can use both the terms interpretable and explainable interchangeably. Despite unbelievable breakthroughs of deep neural networks, they are notorious for the ambiguity in the decision-making process. While powerful, deep learning models are difficult to interpret, and thus often treated
as a black-box. Hence, people are reluctant when considering deep learning in critical applications, such as disease prediction or autonomous driving. An answer to unveil the myth of deep neural network is Interpretable Machine Learning (or Interpretable ML), which helps users to understand neural networks from inside as in Fig. \ref{fig:black-box}. More precisely, contemporary interpretability methods bring us advantages to understand the decision-making process of deep neural networks \cite{simonyan2013deep, dabkowski2017real, nguyen2021effectiveness}.

\subsection{Network Dissection}
\begin{figure*}[h!]
	\centering
	\includegraphics[scale=0.35]{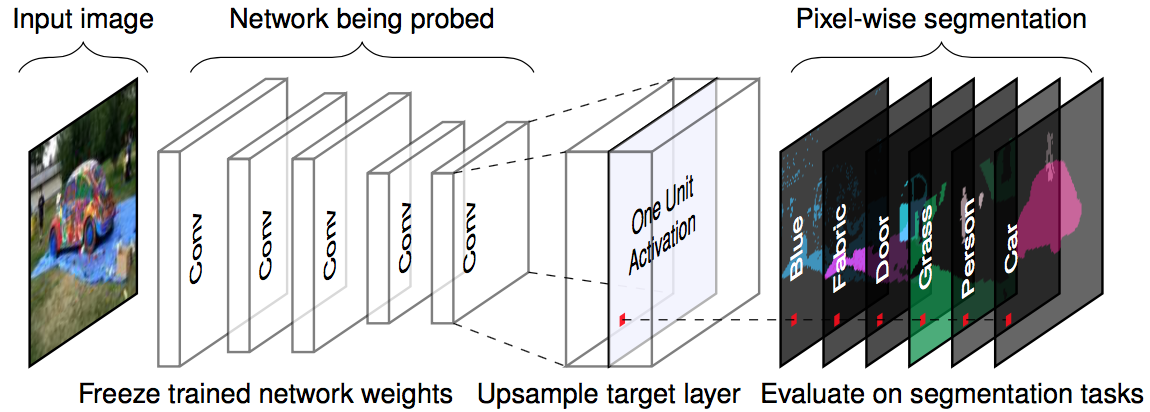}
	\caption{Network dissection algorithm workflow.}
	\label{fig:dissection}
\end{figure*}

Network dissection categorizes network components (e.g. channels) as human concepts. This line of research is based on a hypothesis that a neural network learns disentangled concepts. For instance, channel 241 detects red color, and channel 999 is interested in human eyes. Bau and Zhou et al. \cite{bau_zhou} use a trained model and propagate an image up to a target layer, then upscale the activations to match the original image size and compare with the segmentation ground-truth (shown in Fig. \ref{fig:dissection}).

There are two notable findings that motivate us to study catastrophic forgetting in neural networks. Many channels detect the same concept and in transfer learning, the role of a given channel can totally change. For example, a cat detector becomes a wheel detector. 

\subsection{Feature Visualization}
\begin{figure*}[h!]
	\centering
	\includegraphics[scale=0.45]{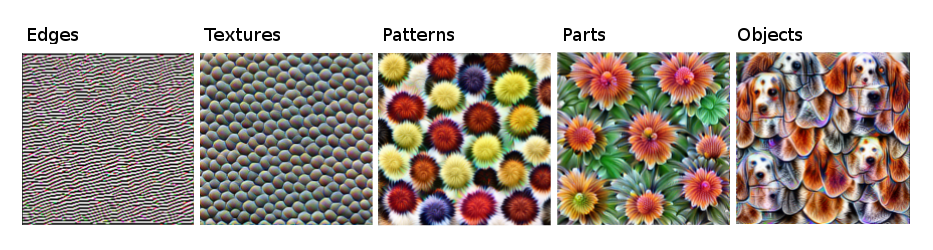}
	\caption{Abstract features in hidden layers of neural networks.}
	\label{fig:features}
\end{figure*}

Abstract features and concepts are learned by a network during training; and thus, in feature visualization approaches, being visualized via activation maximization. One of the greatest strengths of neural networks is that they absorb high-level features in hidden layers, wiping out the need for traditional feature engineering. The deeper a layer is, deep more sophisticated features it captures. As shown in Fig. \ref{fig:features}, the complexity of learned features is increasing from left to right, indicating the depth of layers.

\begin{figure*}[h!]
	\centering
	\includegraphics[scale=0.1]{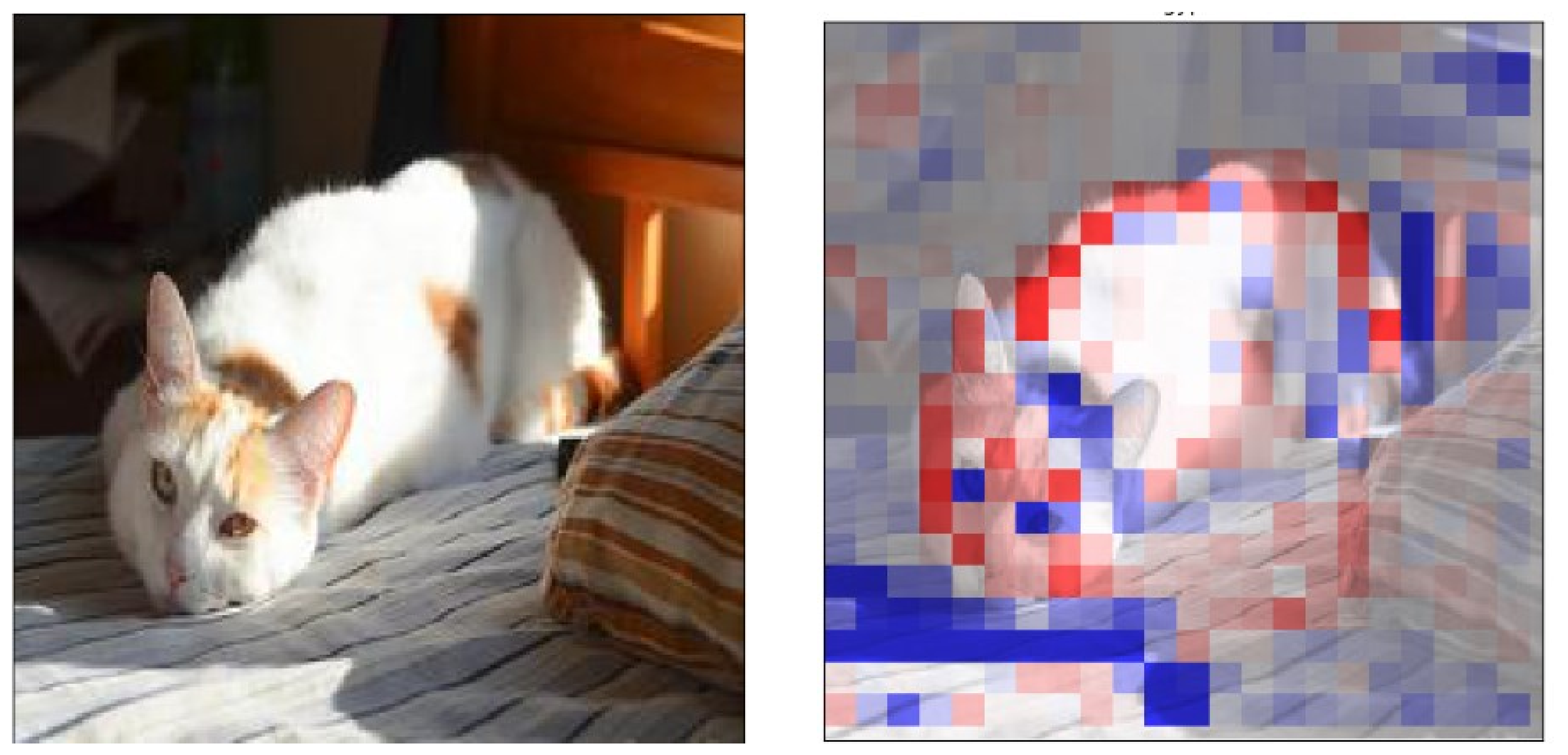}
	\caption{PDA highlights pixels that are evidence for (red) and against (blue) a prediction of cat.}
	\label{fig:pda-cat}
\end{figure*}

In \cite{zeiler2014visualizing}, deconvnets allow us to recognize which features are expected by a specific part of a network or what properties of image excite a chosen neuron the most.
In stark contrast to feeding an input image to diagnose, Yosinski et al. \cite{yosinski2015understanding} attempt to generate an image which maximizes the activation of a given neuron by gradient descent algorithm.
Visualizing the activation of a neuron or a layer in networks helps us categorize the specific role of each block, layer, or even a node. It has been proved that the earlier layers extract local features, such as edges or colors; while deeper layers are responsible for detecting globally distinctive characteristics. Prediction Difference Analysis (PDA) \cite{zintgraf2017visualizing} - shown in Fig. \ref{fig:pda-cat}, even more specifically, highlights pixels that support or counteract a certain class, indicating which features are positive or negative to a prediction.  

However, these tools only provide the computer vision, leaving the conclusion for users. This manual process can not ensure the quality of the observation when we may have hundreds or even thousands of feature maps. \texttt{CFD} automatically detects the forgetting components in a network, entirely leveraging the generated feature maps from PDA \cite{zintgraf2017visualizing}.

\section{Continual Learning}
Continual learning requires an ability to learn over time without witnessing catastrophic forgetting (known as semantic drift), and allows neural networks to incrementally solve new tasks. To achieve these goals of continual learning, studies mainly focus on addressing catastrophic forgetting problem. While the community has not yet agreed on a shared categorization for continual learning strategies, Maltoni and Lomonaco \cite{maltoni2019continuous} propose a three-way fuzzy categorization of the most common continual learning strategies: architecture, regularization, and rehearsal.

Rehearsal approaches \cite{lopez2017gradient, tasar2018incremental, castro2018end} demand data from old tasks, leading to privacy or limited storage budget issues, making them non-scalable if tasks arrive ceaselessly. Architectural techniques \cite{rusu2016progressive, lopez2017gradient, yoon2017lifelong, parisi2018lifelong}, at the same time, alter the original network architecture for adapting to new tasks besides preserving old memory. When a new task comes, layers or neurons are typically added to the previous model. Consequently, the capacity of the network is increased accordingly. This leads to a storage problem when the number of new classes is numerous because the final model could be gigantic, alleviating its portability. The concept of regularization approaches \cite{zenke2017continual, kirkpatrick2017overcoming, li2017learning} is to keep the model close to its previous state by adding a penalty in the objective function. Neither accessing old samples nor expanding the old network is required in regularization methods. 

Many works have managed to address the forgetting problem in generative models \cite{zhai2019lifelong}, image captioning \cite{nguyen2021explaining, nguyen2020dissecting, nguyen2019contcap_scalable}, or semantic segmentation \cite{tasar2019incremental}. Toy datasets are preferred to be used when proving the feasibility of continual learning algorithms. Necessarily, standard datasets are tweaked to fit the experimental setup. Particularly, continual learning has been experimented on benchmarks like MNIST variants \cite{kirkpatrick2017overcoming, lee2017overcoming, nguyen2017variational, yoon2017lifelong}, CIFAR10/100 Split \cite{zenke2017continual}, and ImageNet ILSVRC \cite{russakovsky2015imagenet} for image classification. Shmelkov et al. \cite{shmelkov2017incremental} use PASCAL VOC2007 \cite{everingham2010pascal} and COCO datasets \cite{lin2014microsoft} for object detection; and PASCAL VOC2012 \cite{pascal-voc-2012} is selected in \cite{michieli2019incremental} for semantic segmentation. Seff et al. \cite{seff2017continual} suggest to use an EWC-augmented loss \cite{kirkpatrick2017overcoming} to counteract catastrophic forgetting in synthetic image generation task while Lifelong GAN \cite{zhai2019lifelong} leverages knowledge distillation to transfer knowledge to the new network. 

A work from Kemker et al. \cite{kemker2018measuring} conducts experiments on state-of-the-art continual learning techniques that address catastrophic forgetting. It is demonstrated that although the algorithms work, but only on weak constrains and unfair baselines, thus the forgetting problem is not fully solved and witnessed yet. They insist on the infeasibility of using toy datasets, such as MNIST \cite{kirkpatrick2017overcoming} or CIFAR \cite{zenke2017continual} in continual learning experiments. As a result, this work inspires us to create a new dataset from MS-COCO 2014 (Split MS-COCO).

\newpage

Fine-tuning is considered as a baseline in \cite{shmelkov2017incremental, tasar2019incremental, michieli2019incremental, rebuffi2017icarl}. Freezing either a few specific layers or a major part of a network is proposed in \cite{michieli2019incremental}, which reveals that just simply keeping some parameters unchanged can greatly helps networks become robust against catastrophic forgetting. Learning without Forgetting (LwF) \cite{li2017learning} utilizes old models to generate pseudo data, which guides the new model to reach a shared low-error region of problems. Also leveraging the old networks, knowledge distillation approaches \cite{michieli2019incremental, hou2018lifelong} are formally recognized to facilitate better generalization in life-long learning via a teacher-student learning strategy. Nevertheless, algorithms are inclined to rely on external factors (e.g., input or rehearsal data, objective functions \cite{li2017learning, michieli2019incremental}) while ignoring the question of why catastrophic forgetting internally happens. In this research, our algorithm is derived from catastrophic forgetting exploration.

\newpage

\chapter{Approach}
\label{sec:approach}
Although research from \cite{nguyen2019toward} shows an interest in understanding catastrophic forgetting, they focus on how task properties influence the hardness of sequential learning. Hence, they are explaining based on the input data. \texttt{CFD} approaches the problem from an alternative perspective, trying to explain how forgetting happens over time based on the computer vision of models. In comparison, the ultimate goal of this tool is to figure out the most plastic layers or blocks in a network. Plasticity means a low degree of stiffness or being easy to change. Although a variety of continual learning techniques have been proposed to alleviate catastrophic forgetting, none of them takes advantage of the findings from Interpretable ML. Critical freezing is built on the top of \texttt{CFD}'s investigation to provide an interpretable and effective approach to deal with catastrophic forgetting. In the learning process, the optimal state of the old model is employed to initialize the new network. This way mimics the working mechanism of the human brain.

Derived from the network structure, we propose to freeze the encoder and decoder in turns (partially freezing) to moderate semantic drift. Pseudo-labeling makes use of the previous model to interpret predictions of the previous model on the coming samples while knowledge distillation utilizes the previous model as a teacher model, forcing a student to not forget solved problems.

\section{Partially Freezing for Image Captioning}
A heuristic solution to hinder catastrophic forgetting is to freeze a part of model. As our architecture is divided into an encoder and a decoder, keeping one component intact can affect directly to the performance on the previous tasks as well as the new task. 

\begin{figure*}[h!]
	\centering
	\includegraphics[scale=0.45]{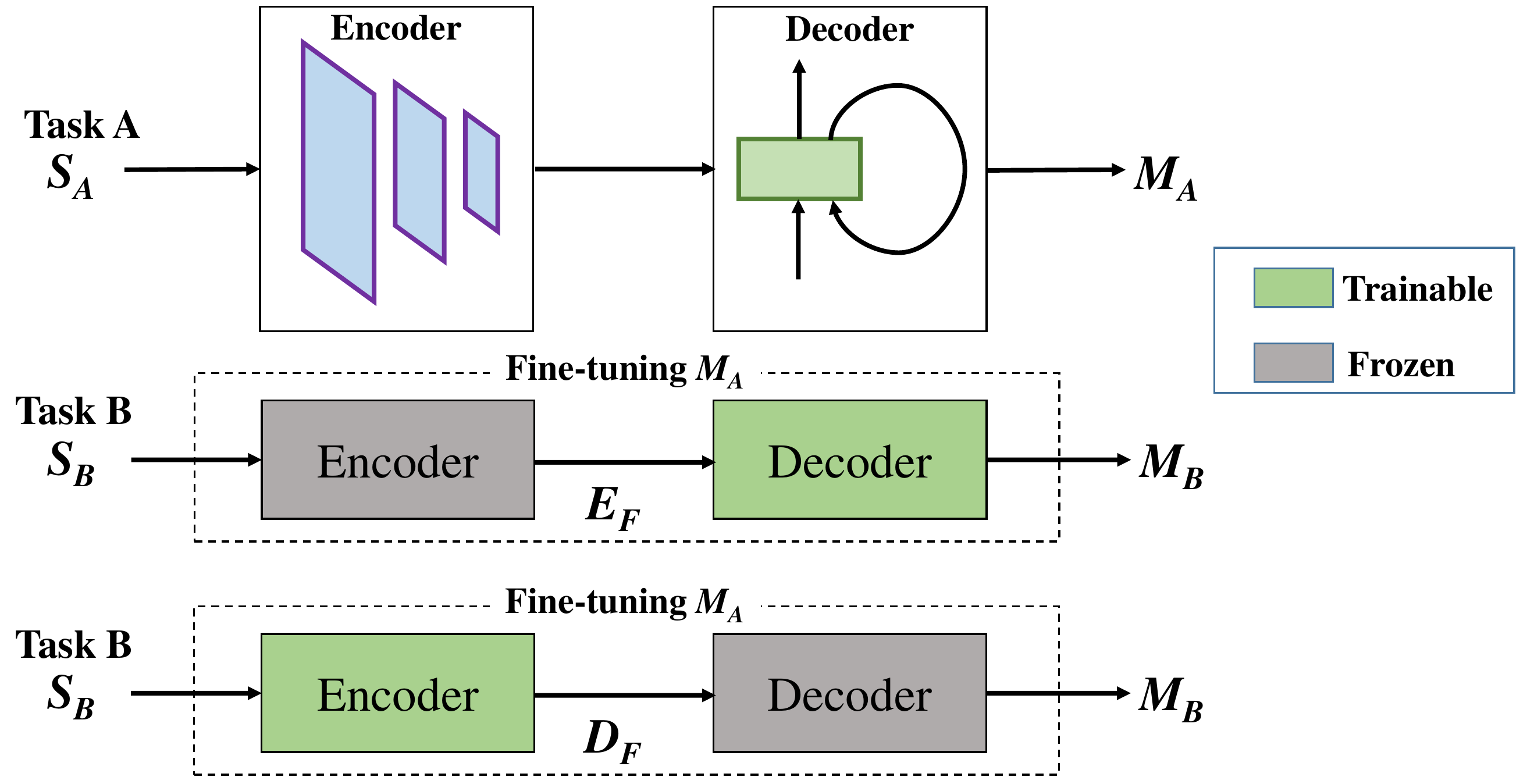}
	\caption{Freezing a component of the fine-tuned model while the remaining component is trainable.}
	\label{fig:freezing}
\end{figure*}

On the one hand, the encoder is frozen and the decoder is updated exploiting the new dataset. The feature extraction remains compared to the previous model $M_{A}$ which operates as a fine-tuned model (Task A). The feature extractor $E$ is constrained, leaving combatting with the new task for the decoder $D$, we refer as $E_{F}$ (see Fig. \ref{fig:freezing}). The vocabulary is expanded to ensure the naturalness of the prediction, and the decoder is modified according to the vocabulary size. 

On the other hand, we freeze the decoder $D$ while the encoder is trainable ($D_{F}$ as in Fig. \ref{fig:freezing}). Because neurons are added to the decoder as by virtue of expanding vocabulary, the newly attached neurons are the only trainable part of the decoder. By setting like this, the convolutional network can better learn unseen features, thus feeding more fine-grained input into the decoder. In the case of both $E$ and $D$ being frozen, the new model is the same as the previous model, while fine-tuning is when both $E$ and $D$ are trainable.

The objective function when applying freezing techniques is the standard loss function for image captioning:

\begin{equation}
\label{eq:0}
\begin{aligned}
	\mathcal{L} ={} & L_{CE}\\
	            ={} & - \sum_{i=1}^{v} Y_{k}^{i} \log \hat{Y}_{k}^i
\end{aligned}
\end{equation}
In the equation \eqref{eq:0}, $L_{CE}$ is the cross-entropy loss over annotation and the prediction, $v$ is the size of vocabulary. $Y_{k}$ is the ground truth (one-hot vector), and $\hat{Y}_{k}$ is the predicted caption (probability distribution over vocabulary). 

\section{Pseudo-labeling (Learning without Forgeting - LwF)}
We refer this method as pseudo-labeling since when training, pseudo-labels are generated to guide the current model to mimic the behavior of the previously trained model. The procedure of this approach is described as follows:

\begin{lstlisting}[mathescape,escapechar=\%]
At task $k^{th}$, 
%\underline{\textit{Start with}}%:
  $\theta_{k-1}$ %\space\space\space: parameters of the old task%
  $X_k$, $Y_k$ %: training images and reference captions on the new task%
%\underline{\textit{Initialize}}%:
  $Y_{k-1}$ $\leftarrow$ $M(X_{k},\theta_{k-1})$ %\space// infer captions by the old%
                        // model on new data
  $\theta_k \leftarrow \theta_{k-1}$ %\space\space\space\space\space\space\space\space\space\space\space\space\space\space\space// initialize new model by old%
                        // parameters
%\underline{\textit{Train}}%:
  $\hat{Y}_k \leftarrow M(X_k,\hat{\theta}_k)$ %\space\space\space\space\space\space// new task prediction%
  $\theta_k^{*} \leftarrow \underset{\hat{\theta}_k}{argmin}(L(Y_k,\hat{Y}_k) + L(Y_{k-1},\hat{Y}_k))$
\end{lstlisting}

In \cite{li2017learning}, they use task-specific parameters to generate the pseudo labels while in our proposed pseudo-labeling approach, we use a unified network for the whole process. This point eases the training because we do not need to separate the network into sub-modules. We also do not apply knowledge distillation in pseudo-labeling approach because we do not expect the logit values in captioning task. The pseudo labels are fake captions on the new data.

Pseudo labels $Y_{k-1}$ are acquired by inferring the caption of all input images in the dataset of the new task using the previous model. Although new classes appear, the convolutional network architecture stays unchanged during training since the expected output is textual caption, not the probabilities for classes as in object classification or semantic segmentation. After being initialized by the foregoing model, we run training in a supervised manner to minimize the loss computed based on the ground truth, the predicted caption, and the pseudo labels $Y_{k-1}$. The loss component from pseudo-labeling $L_{P}$ is explicitly written as:

\begin{equation}
\begin{aligned}
    {L_{P}} = - \beta \cdot \sum_{i=1}^{v} Y_{k-1}^i \log \hat{Y}_{k}^i
\end{aligned}
\end{equation}
where $\beta$ is considered to be a regulator to accentuate the old or new task ($\beta = 1$ in experiments), and the final objective function for this approach is:
\begin{equation}
\begin{aligned}
	\mathcal{L} ={} & L_{CE} + L_{P}\\
	            ={} & - \sum_{i=1}^{v} Y_{k}^{i} \log \hat{Y}_{k}^i - \beta \cdot \sum_{i=1}^{v}Y_{k-1}^i \log \hat{Y}_{k}^i
\end{aligned}
\end{equation}

\section{Knowledge Distillation}
This technique frames the training process in a teacher-student strategy when we want to deliver the knowledge acquired from one model to another model. By applying distillation, teacher is the model obtained from the old task while student is the new model. Once an image $X_{n}$ is passed through both teacher and student, the mean squared error between outputs of student and teacher ($Y_{k}^{st}$ and $Y_{k}^{tr}$) is added to the loss function of student to penalize it. We propose to distill the feature learned by the encoder to provide the student decoder the flexibility adapting to the new task. This approach is felicitous because the spatial information is global while the sentence semantics changes constantly over domains. Thus, the decoder is freer to change, in comparison with the encoder. The distillation term $L_{dis}$ is computed as:
\begin{equation}
\begin{aligned}
	{L_{dis}} = \lambda\cdot\left\| Y_{k}^{tr} - Y_{k}^{st} \right\|^2 
\end{aligned}
\end{equation}
where $\lambda$ may be increased to encourage student to imitate the behaviors of teacher more intensively ($\lambda = 1$ in experiments). The loss function is:
\begin{equation}
\begin{aligned}
	\mathcal{L} ={} & L_{CE} + L_{dis}\\
	            ={} & - \sum_{i=1}^{v} Y_{k}^{i} \log \hat{Y}_{k}^{i}  + \lambda\cdot\left\| Y_{k}^{tr} - Y_{k}^{st} \right\|^2 
\end{aligned}
\end{equation}

\section{Catastrophic Forgetting Dissector - CFD}

\begin{figure*}[ht!]
\begin{center}
\includegraphics[scale=0.35]{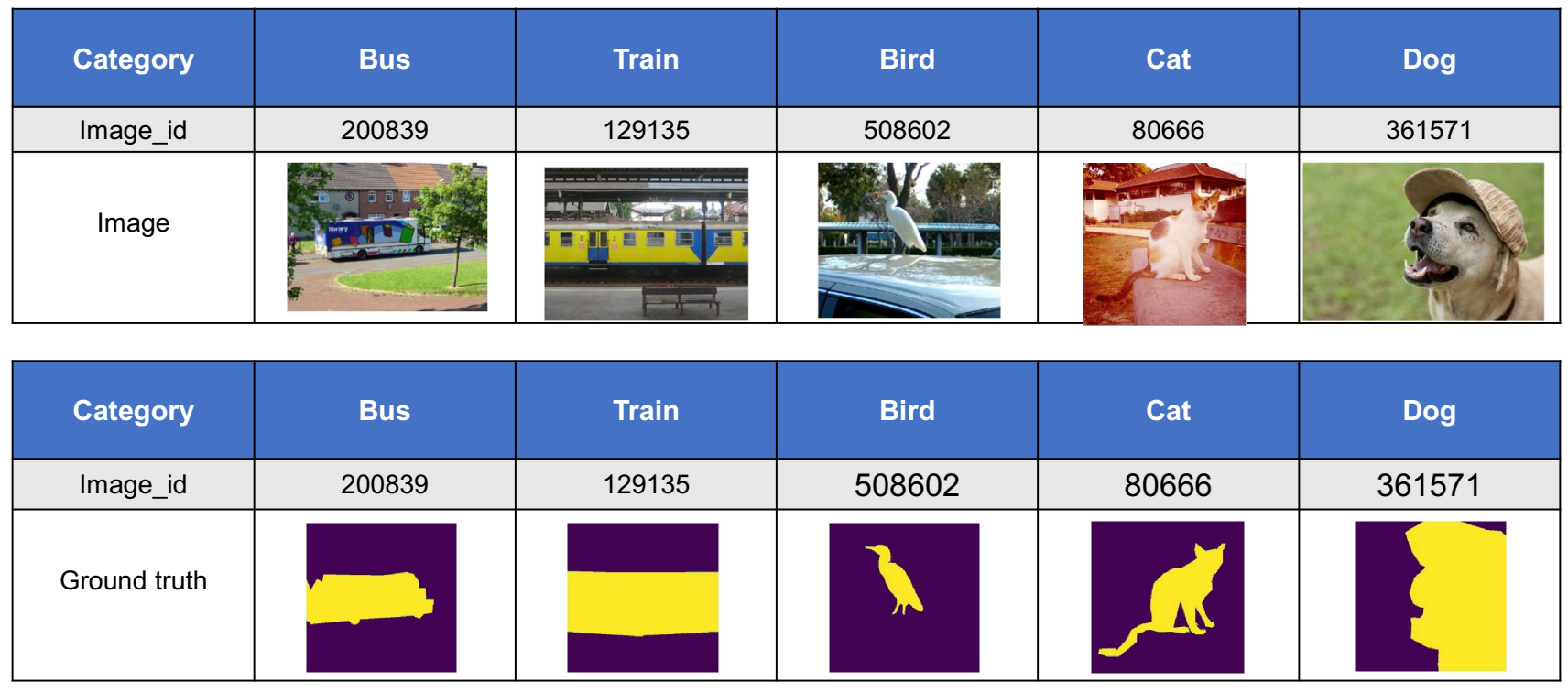}
\end{center}
   \caption{Segmentation ground truth as input of \texttt{CFD}.}
\label{fig:input}
\end{figure*}

To dissect the model, we use the activation of hidden layers like \cite{tran2021simple} but to understand the forgetting effect. We initially hypothesized that different features of the objects could be captured and visualized by particular feature maps in different layers. By looking into each response map in one convolutional block (conv block), we realize diverse features, such as eyes, face shape, car wheels, or background are isolatedly recognized by different channels, which avers our hypothesis. Unfortunately, finding each feature map manually by human eyes might be inefficient. To solve this issue, we only compare the computer vision with the ground truth segmentation rather than small details. The ground truth is sampled as in Fig. \ref{fig:input}. In general, we do not seek for the answer that what features are being forgotten, but which layers are now forgetting.

\begin{figure*}[hb!]
	\centering
	\includegraphics[scale=0.6]{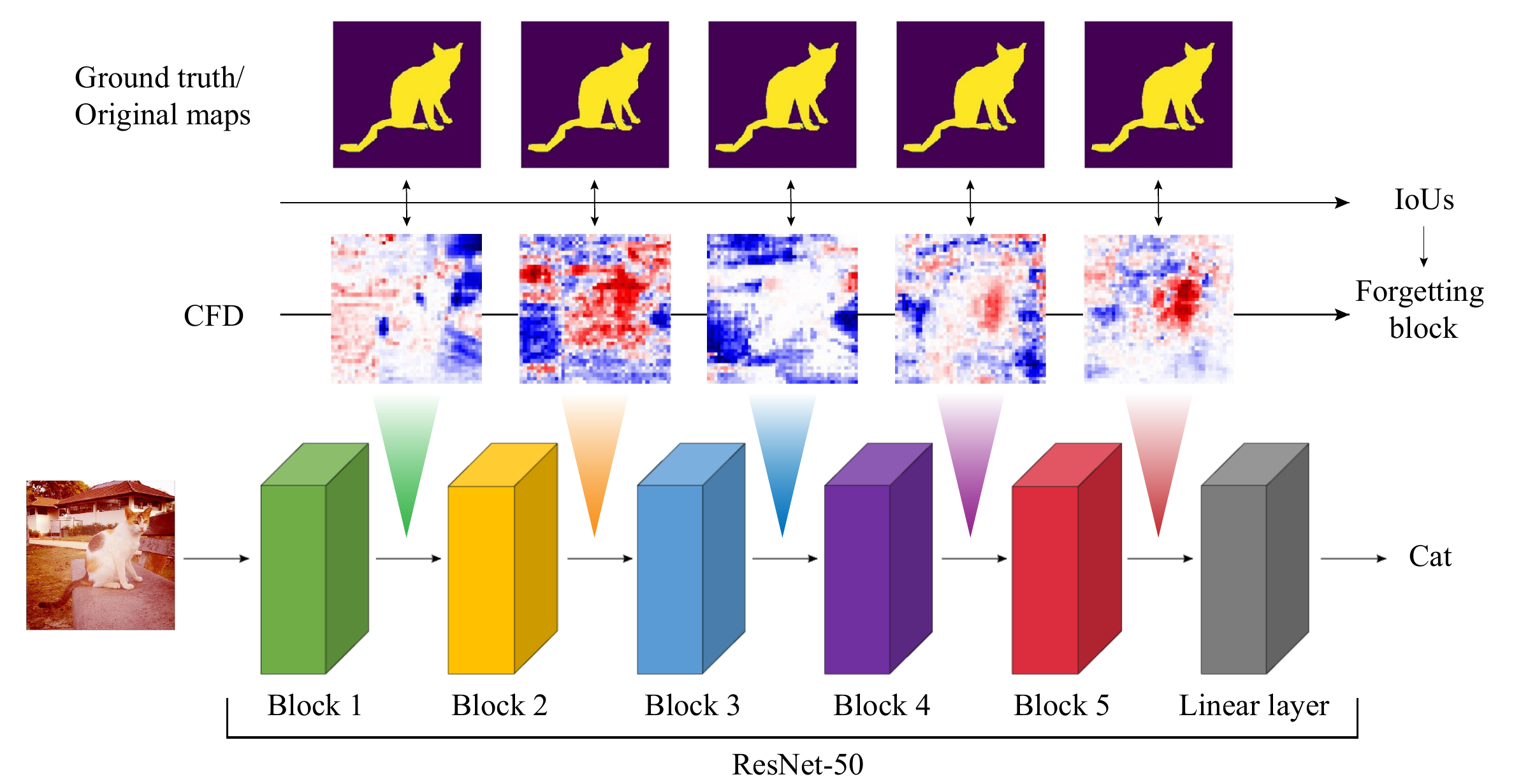}
	\caption{\texttt{CFD} working on ResNet-50 that includes 5 convolutional blocks.}
	\label{fig:CFD}
\end{figure*}

The scheme of \texttt{CFD} is illustrated in Fig. \ref{fig:CFD}. The visualization from the tool and the ground truth segmentation are simultaneously employed to infer the forgetting blocks. The visualizations of the tool and the ground truth segmentation are simultaneously employed to infer the forgetting blocks. We assume that the semantic segmentation label of MS-COCO dataset \cite{lin2014microsoft} is what human eyes perceive. Next, we compare this segmentation with the computer vision of the model, particularly concentrating on positive evidence for a prediction to see how supportive features are disregarded. 

\begin{figure*}[hbt!]
\begin{center}
\includegraphics[scale=0.28]{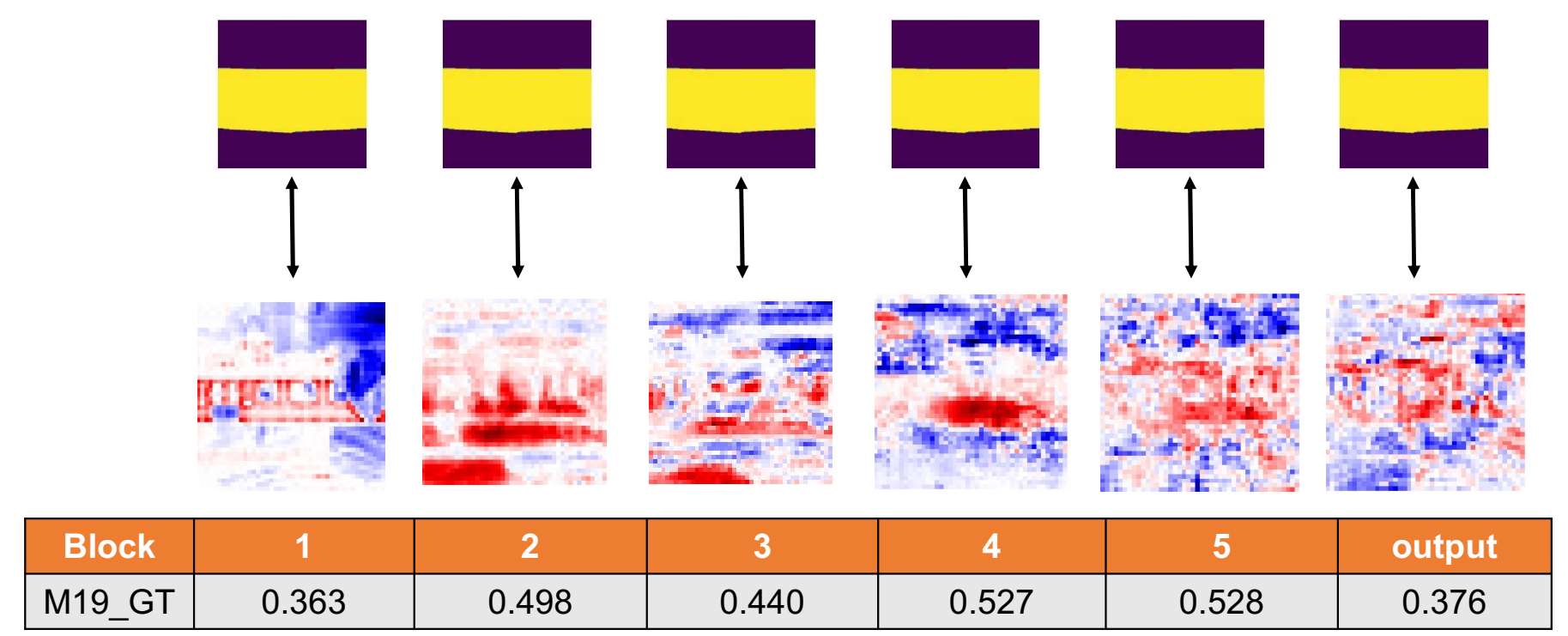}
\end{center}
   \caption{Choosing representatives based on overlap between feature maps and ground truth.}
\label{fig:iou_1}
\end{figure*}

\begin{figure*}[hbt!]
\begin{tabular}{cc}
\bmvaHangBox{\includegraphics[width=6.7cm]{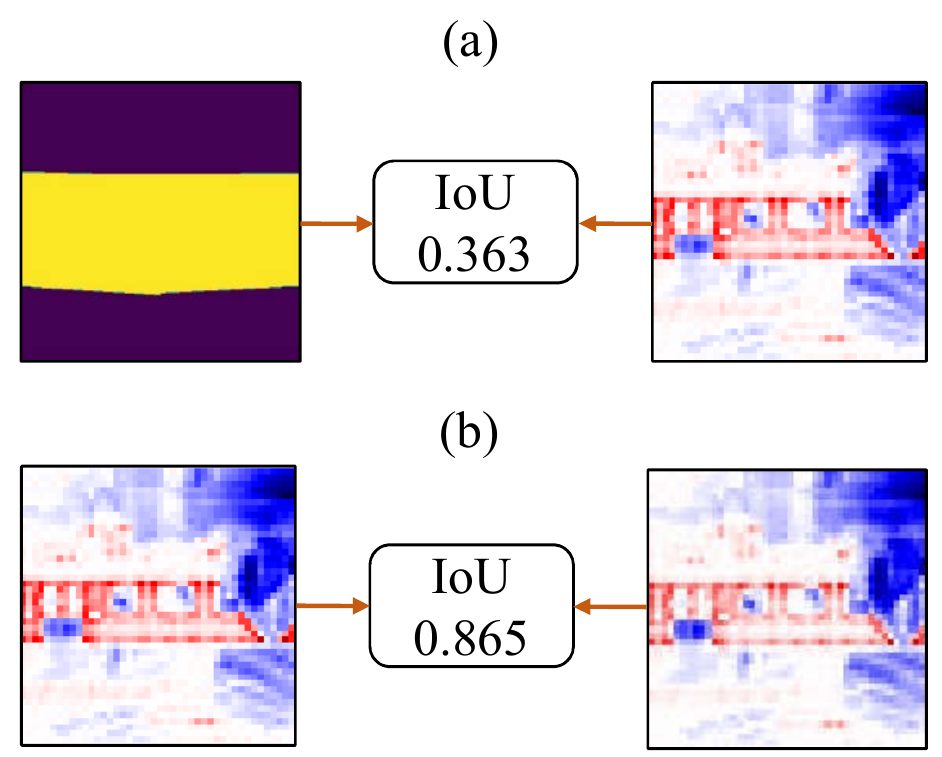}}&
\bmvaHangBox{\includegraphics[width=8.5cm]{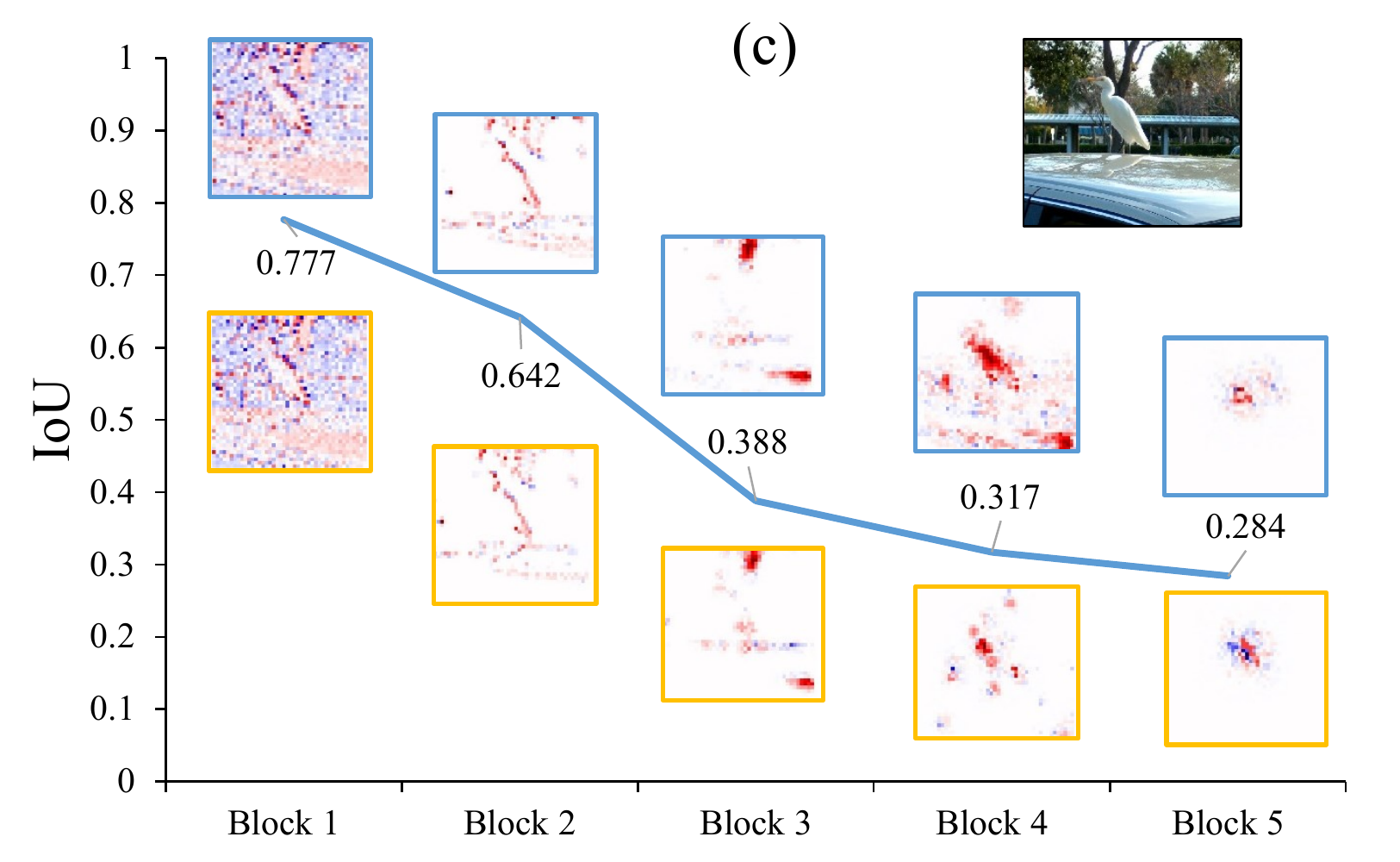}}\\
\end{tabular}
   \caption{(a) IoU value between the segmentation of a train and the positive features. (b) The IoU on the positive features of two representative maps. Red is evidence and blue is against. (c) IoU graph and computer vision of a bird image. Yellow-bounding and blue-bounding boxes are from the old and new model respectively.}
\label{fig:IoU1}
\end{figure*}

The IoU value between the segmentation and evidence is calculated as shown in Fig. \ref{fig:IoU1} (a). On the right side of Fig. \ref{fig:IoU1} (a), we have an input image of a train, red dots advocate the fact that the output should be ``\textit{train}'' while blue ones contradict this prediction.

Having the \textit{m-th} feature map ($FM$) in the \textit{l-th} layer of a model \textit{M} and the ground truth segmentation \textit{GT}, the IoU is computed as in equation \eqref{eq:1}:
\begin{equation}
\label{eq:1}
IoU_{M,GT} (l,m) =  \frac{FM(l,m) \cap GT }{FM(l,m) \cup GT}
\end{equation}

To select the feature map having the largest overlap with the ground truth in \textit{l-th} conv block, the representative feature map (RM) with the best IoU is $RM_{M,GT}$ :
\begin{equation}
\label{eq:2}
RM_{M,GT}  (l) = argmax_m (IoU_{M,GT} (l,m))
\end{equation}

To understand how the computer vision changes over the training process, we compare the $RM$s in the new model with the $RM$s of the old model shown in Fig. \ref{fig:IoU1} (b). We compute $IoU_{M_O,GT}$ by equation \eqref{eq:1} then achieve $RM_{M_O,GT}$ from equation \eqref{eq:2} ($RM_{M_O,GT}$ is the representative feature map of the old model). The forgetting effect of each trained model is measured by the IoUs between $RM_{M_O,GT}$ and feature maps of  \textit{$M_N$} ($M_N$ is the new model):
\begin{equation}
\label{eq:3}
IoU_{M_N,M_O} (l,m) = \frac{F M(m) \cap RM_{M_O,GT} }{F M(m) \cup RM_{M_O,GT}}
\end{equation}

\newpage

Similar to the method of finding out the best map fitting with ground truth, the feature map representing the best memory of the original feature map is denoted as $RM_{M_N,M_O}$ in equation \eqref{eq:4}. $RM_{M_N,M_O}$ is specified as: 
    
\begin{equation}
\label{eq:4}
RM_{M_N,M_O}  (l) = argmax_m (IoU_{M_N,M_O} (l,m))
\end{equation}

\begin{figure*}[hbt!]
\begin{center}
\includegraphics[scale=0.3]{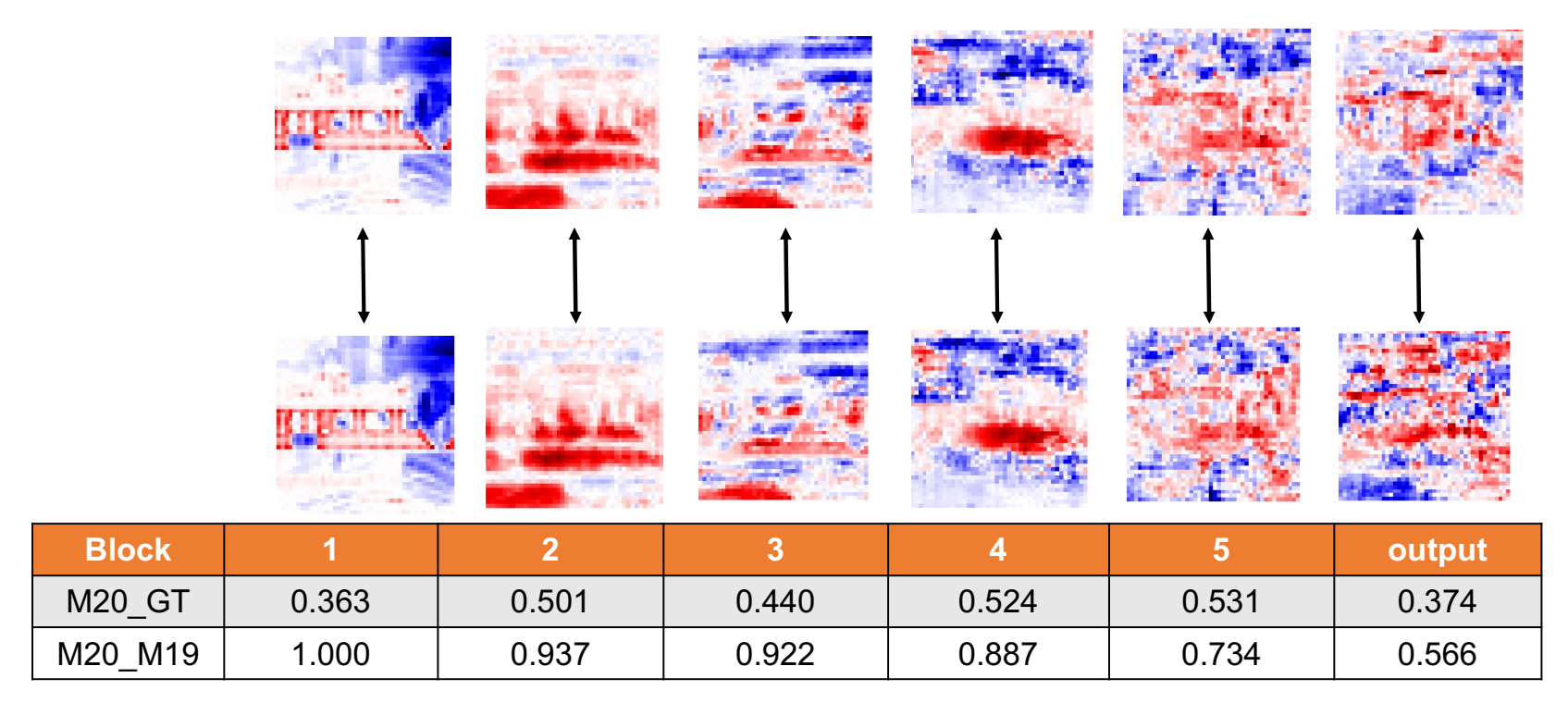}
\end{center}
   \caption{Choosing representatives based on computer visions of two models.}
\label{fig:iou_2}
\end{figure*}

For instance, when a new class is added, we compare the computer vision of the consecutive models to determine which feature maps generated by the new model should be used to observe the forgetting. As shown in Fig. \ref{fig:iou_2}, the representative on the first layer on the new model should have the largest overlap in comparison to the representative on the same layer of the old model. The procedure is done similarly on other layers.

In the same block of both the old and new model, the role of a filter can be adjusted. For instance, the $50^{th}$ filter in the $2^{nd}$ block of the old model detects the eyes, but the same filter in the same block of the new model may consider the face. Hence, we should not make a comparison based on the index of a filter.

\begin{algorithm}[bt]
   \caption{CFD}
   \label{alg:cfd}
\begin{algorithmic}[1]
   \STATE {\bfseries Input:} Sample set $S$, segmentation ground truth $GT$, old model $M_O$, new model $M_N$, number of blocks $K$ 
   \STATE {\bfseries Output:} Forgetting layer $\mathbb{F}$
   \STATE $i=0$
   \STATE $\L = \emptyset$
   \REPEAT
   \STATE $I = S[i]$
   \STATE $IoUs = \emptyset$
   \STATE $FM = PDA(I)$
   \FOR{$j=1$ {\bfseries to} $K$}
   \STATE $RM_{M_O,GT} \leftarrow \textit{FM with highest } {IoU_{M_O,GT}}$
   \STATE $RM_{M_N,M_O} \leftarrow \textit{FM with highest } {IoU_{M_N,M_O}}$
   \STATE \textbf{Append}($IoUs, max(IoU_{M_N,M_O})$)
   \ENDFOR
   \STATE $\boldsymbol{b} \leftarrow \textit{blocks with the highest drop in IoUs}$ 
   \STATE \textbf{Append}(\L, $\boldsymbol{b}$)
   \STATE $i = i+1$
   \UNTIL{$i =  \textbf{size}($S$)$}
   \STATE $\mathbb{F}$ $\leftarrow$ \textit{Most frequent block in} \L
\end{algorithmic}
\end{algorithm}
    
The workflow of \texttt{CFD} is given by Algorithm \ref{alg:cfd}. The sample set $S$ is described in Section \ref{sec:exp}, $GT$ is the segmentation ground-truth from MS-COCO dataset, $M_{O}$ and $M_{N}$ are the old and new model for comparison respectively, and $K$ is the number of the conv blocks in the network ($K=5$ with ResNets). $\L$ is a list containing the most forgetting layer with respect to all the images in $S$. By inputting an image $I$ from $S$, we get the visualization of feature maps ($FM$) over $M_{O}$ and $M_{N}$ by PDA. However, we need to pick the representative feature map (RM) amongst thousands of feature maps in a conv block.

To choose the representative feature map $RM$ of the $j^{th}$ conv block in a model, we define representative feature map $RM$ to be the feature map having the largest overlap with the $RM$ of the previous model ($RM_{prev}$ in short) for the same $j^{th}$ block. Particularly, the $RM_{prev}$ of $M_{O}$ is the ground truth because $M_{O}$ is the starting model, and $RM_{M_O,GT}$ is the representative feature map at $j^{th}$ block of $M_{O}$. Likewise, $RM_{prev}$ of $M_{N}$ at $j^{th}$ block is $RM_{M_O,GT}$, and we obtain $RM_{M_N,M_O}$ by comparing feature maps of $M_{N}$ and $RM_{M_O,GT}$. The IoU values between $RM_{M_O,GT}$ and $RM_{M_N,M_O}$ are calculated at each conv block and appended to a list $IoUs$. After calculating IoU drops through the ResNet blocks and denote the block giving the highest drop as $b$, we can put $b$ as the block where the most substantial forgetting happens into a list $\L$, tested on the input image $I$. Finally, we generalize on all images of $S$ to return the most forgetting component $\mathbb{F}$. 

In the juxtaposition of the old and new model, the IoUs are visually drawn to provide a bird-eye view of the forgetting trend shown in Fig. \ref{fig:IoU1} (c). We may argue that the conv block having the lowest IoU (block 5) is the victim of catastrophic forgetting. However, this assumption is not asserted because of the error accumulation in deep neural networks. The computer vision of deeper layers is directly attributed to earlier conv blocks. Once forgetting occurs in the first layer, it will be propagated throughout the entire network. 

We propose to leverage IoU slopes to find the weakest block $b$. At a point, if the IoU drops significantly compared to the previous value, it should be the sign of catastrophic forgetting. In Fig. \ref{fig:IoU1} (c), a plummet of the IoU value is seen between block 2 and block 3 (0.642 to 0.388). The first thought appearing in our mind was that the $3^{rd}$ is forgetting most catastrophically. It is true but we need to regard the fact that the worst map in block 3 is a result of block 2 and block 1. This finding is the foundation for our technique to prevent catastrophic forgetting. 

\section{Critical Freezing}
Fine-tuning techniques play a pivotal role in training deep networks if data distribution evolves. Regarding a pre-trained model, the feature extractor which captures global information is carefully protected in adaptation. The output layer may be superseded, or the learning rate should be tweaked to a tiny number. Another dominant approach is to freeze the weights of the early layers. They are all effective yet ambiguous because we can not ensure freezing which layers will give the best result. Using the investigation from \texttt{CFD}, we freeze the precursors of the most plastic layer in a deep neural network. If a network has $K$ conv blocks, and we find the $F^{th}$ convolutional layer broken, then we try to freeze earlier layers than the $F^{th}$ layer. If updating the fragile components is necessary, a learning rate on those blocks should be thoroughly calibrated. The procedure of critical freezing is shown in Algorithm \ref{alg:cf}. The objective function is the standard cross-entropy loss for image captioning, $\mathbb{V}$ is the size of the vocabulary, $Y_{k}^{i}$ is the ground truth, and $\hat{Y}_{k}^i$ is the prediction.

\begin{algorithm}[hbt!]
   \caption{Critical Freezing}
   \label{alg:cf}
\begin{algorithmic}[1]
   \STATE {\bfseries Input:} Sample set $S$, segmentation ground truth $GT$, old model $M_O$, new model $M_N$, number of blocks $K$
   \STATE {\bfseries Output:} optimal state $\theta^{*}$ 
   \STATE $M_N$ $\leftarrow$ $M_O$          // initialize new model by old parameters
   \STATE $F$ $\leftarrow$ \texttt{CFD}($S$, $GT$, $M_O$, $M_N$, $K$)
   
   \FOR{$i=1$ {\bfseries to} $F-1$} 
   \STATE \textbf{grad}($M_N[i]$, $False$) // freeze the layer $i^{th}$
   \ENDFOR
   
   \STATE $\theta^{*}$ $\leftarrow$ $\underset{\hat{\theta}}{argmin}(- \sum_{i=1}^{\mathbb{V}} Y_{k}^{i} \log \hat{Y}_{k}^i)$
\end{algorithmic}
\end{algorithm}

\section{Implementation Details}

\subsection{Feature Extraction}
We use the pre-trained model of ResNet-50 \cite{he2016deep}, removing the final fully connected layer to obtain the immediate features. Behind the ResNet is an embedding layer which has an embedding size of 256. This embedding layer is responsible for bridging the output of the ResNet and the input to the RNN, followed by a batch normalization layer with $momentum = 0.01$.
\subsection{Language Model}
The entrance of the language model is an embedding module to embed the features and labels to feed the LSTM. The RNN contains a single layer LSTM with the size of hidden state is 512 and the embedding size is 256. A fully connected layer is placed at the end of the language model to generate the score for each entry in the vocabulary. Note that when the vocabulary is increasingly accumulated, the embedding layer and the output layer of the decoder are also enlarged accordingly.
\subsection{Training Details}
To help the model work with images of different sizes, we always resize input to square images. We shuffle images at each epoch, and the batch size is 16. Early stopping is used to help models reach the optimal state within 50 epochs of training. The value $patience$ - the number of epochs to wait before early stop if no loss improvement on the validation set, is empirically set to 5. For updating both encoder and decoder, we choose an Adam optimizer \cite{kingma2014adam} with a shared learning rate of $5\times 10^{-4}$. 
Exceptionally, in knowledge distillation, we apply another Adam optimizer to encoder with a learning rate of $5\times 10^{-5}$. The parameters of encoder could be changed dramatically during training time, so we want to slow it down.

All methods but distillation are initialized by the final state of the old task. While joint training is incorporated only in pseudo-labeling, normal training procedure is coupled to each remaining technique. Joint training is when we first freeze the entire model except for the newly added neurons, then train to converge. Subsequently, we unfreeze the model and perform training again until convergence (joint-optimization).

The training environment includes one Nvidia TitanX GPU and an Intel(R) Core(TM) i7 4.0 GHz PC with 16GB RAM, running Ubuntu 16.04 LTS (64-bit).
\newpage
\chapter{Evaluation}
\label{sec:exp}

\begin{figure}[hb!]
	\centering
	\includegraphics[width=1.0\textwidth]{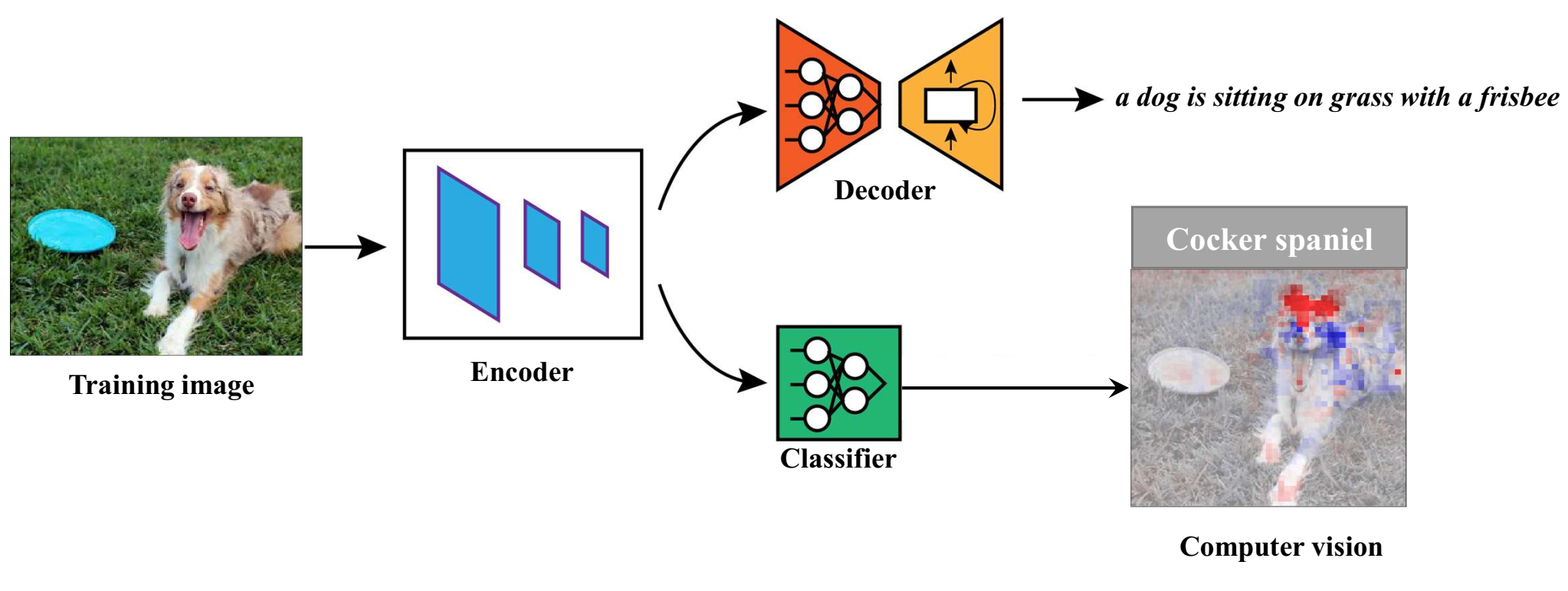}
	\caption{Two-head network for dissecting catastrophic forgetting.}
	\label{fig:two-head}
\end{figure}

The captioning model is divided into an encoder followed by a decoder. Therefore, the original structure should be transformed into a two-head network so that we can get the prediction of the CNN and the sentence generated from the decoder simultaneously. We add one more output layer as a classifier after the encoder, and our proposed architecture is presented in Fig. \ref{fig:two-head}.

\section{Dataset} Our experimental results have been recorded on Split MS-COCO, a variant of the MS-COCO benchmark dataset \cite{lin2014microsoft}. The original MS COCO dataset contains 82,783 images in the training split and 40,504 images in the validation split. Each image of MS COCO dataset belongs to at least one of 80 classes. To create Split MS-COCO, we carry out processing on MS-COCO as the following steps:

\newenvironment{packed_enum}{
\begin{enumerate}
  \setlength{\itemsep}{1pt}
  \setlength{\parskip}{0pt}
  \setlength{\parsep}{0pt}
}{\end{enumerate}}

\begin{packed_enum}
    \item Split all images into distinguished classes. There are 80 classes in total.
    \item Resize images to a fixed size of ($224\times224$) in order to deal with different sizes of input.
    \item Discard images including more than one class in its annotations to obtain ``clear" images only.
\end{packed_enum}

Taking a clear ``dog" image as an example, the reference captions will contain ``dog" in their content and may also have ``grass" (not in 80 classes) but NOT contain any of the 79 remaining classes defined by MS-COCO (cat or frisbee). The reason that we only choose clear images is that we want to assure when facing a new task, objects in images of the new task are unseen. 

Since MS-COCO test set is not available, we divide the validation set of each class into two equal parts: one is for validation and the other for testing. From over 82k images of training set and 40k of validation set from the original MS-COCO 2014, the new dataset has 47,547 images in the training set; 11,722 images in the validation set; and 11,687 images in the test set.

\section{Evaluation Metric} We report the capacity of the framework using traditional scores for image captioning, such as BLEU, ROUGE, CIDEr, and SPICE by \texttt{coco-caption} \cite{chen2015microsoft}. While BLEU and ROUGE are variants of overlapping-based metrics and linearly correlated, CIDEr and SPICE give more weight-age to important terms, thus giving higher correlation with human judgments. For each scenario, we have a table correspondingly. The performance of the techniques on the old task is on the left side while the figure for the new task is located on the right side of the tables.

\section{Addition of Multiple Classes Sequentially}
\begin{figure*}[hbt!]
\begin{center}
\includegraphics[scale=0.15]{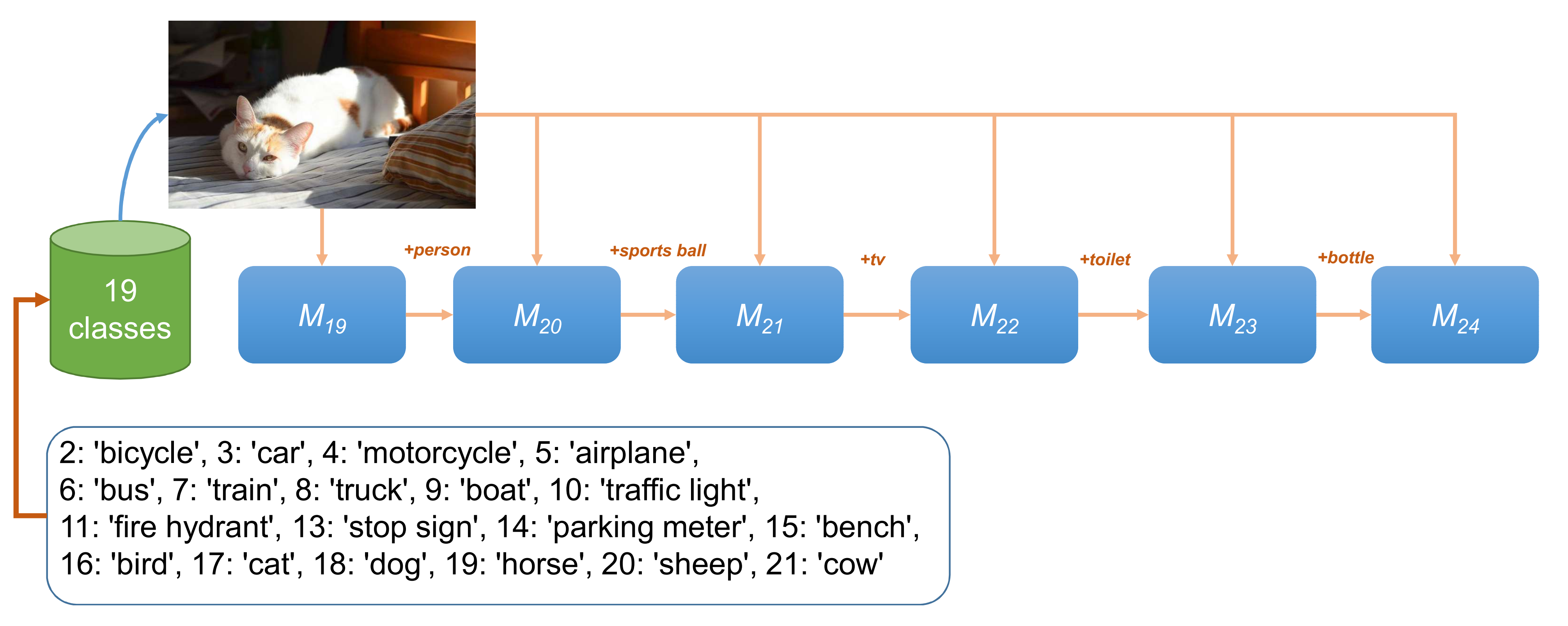}
\end{center}
   \caption{Adding 5 classes one by one scenario.}
\label{fig:scenario}
\end{figure*}

Initially, we train with 19 classes to acquire the base model, followed by adding 5 classes sequentially. The sceraio is illustrated in Fig. \ref{fig:scenario}. 

\begin{figure*}[hbt!]
\begin{center}
\includegraphics[scale=0.4]{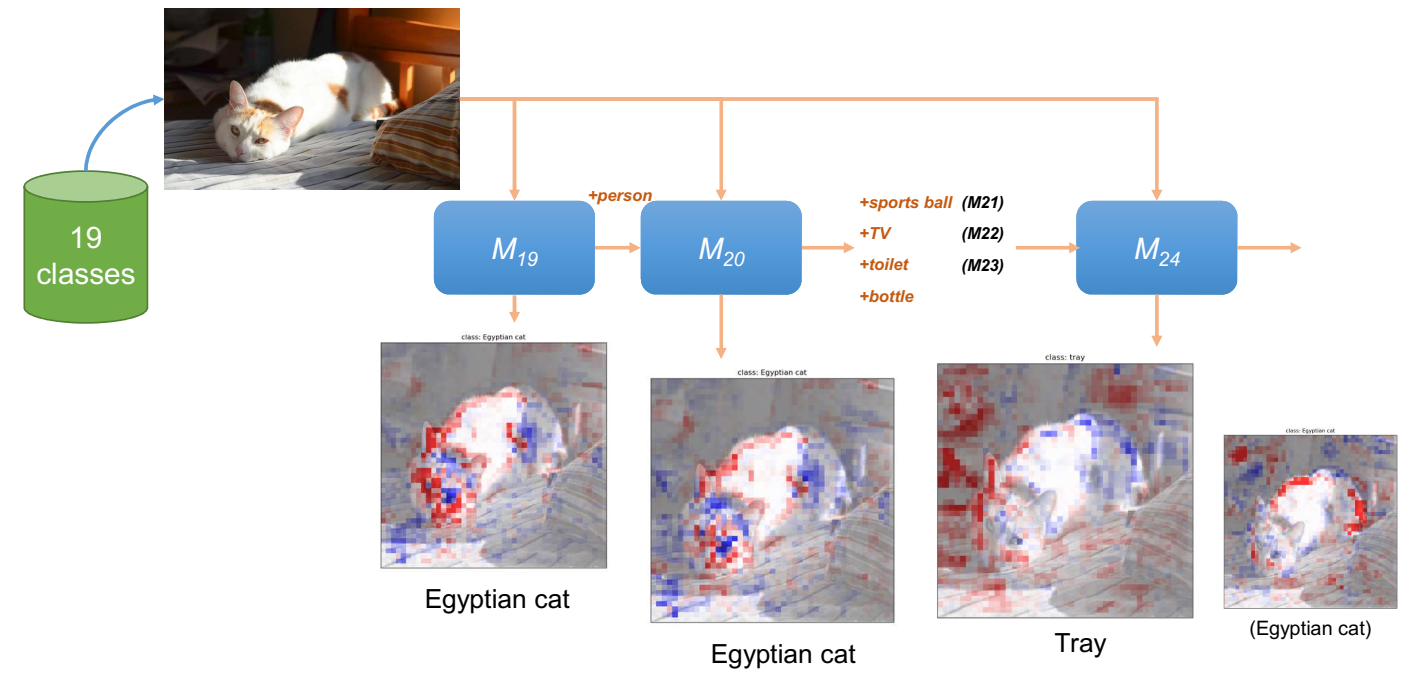}
\end{center}
   \caption{Forgetting in feature extractor when using fine-tuning.}
\label{fig:forgetting}
\end{figure*}

Predictably, when the model faces unseen classes, the knowledge in old classes will be forgotten on the feature extractor. Using the two-head model, we can track both the prediction of the classifier and the computer vision. The model can predict correctly when a class is added, but when the number of classes increases, the prediction becomes totally irrelevant (shown in Fig. \ref{fig:forgetting}).

The captioning model is divided into an encoder and a decoder, in which the encoder is the ResNet-50, and the decoder includes an embedding layer, a single-layer LSTM, and a fully-connected layer producing a word at a time step. As \texttt{CFD} works on a single image, running multiple times on different and diverse input images is needed, helping us to generalize the observation of forgetting. 

We choose a sample set $S$ (\textit{bicycle, car, motorcycle, airplane, bus, train, bird, cat, dog, horse, sheep, and cow}) from 19 trained classes. 

The results of \texttt{CFD} shown in Table. \ref{tab:result_cfd} reinforce the fact that the $3^{rd}$ conv block of ResNet is the most plastic component of this famous conv net given the learning sequence.
To evaluate critical freezing, we perform fine-tuning and various schemes of freezing. In fine-tuning, the old model initializes the new model, and training is done by minimizing the loss on the new task. As the network contains two parts, encoder and decoder, we freeze them separately to specify the best freezing strategy. In addition, we choose various combinations of layers to be frozen besides a famous baseline in continual learning experiments called LwF \cite{li2017learning}. Two knowledge distillation techniques from \cite{michieli2019incremental} are also taken into comparison. The traditional scores for image captioning are considered in evaluating the superiority of critical freezing over the baselines. $BLEU4$ and $ROUGE\_L$ are essentially word-overlap based metrics, while CIDEr and SPICE are more trustworthy because they give more weights on significant terms, such as verbs or nouns. Therefore, we will give discussion only on CIDEr score for conciseness.   

\begin{table*}[hbt]
\begin{center}
\begin{tabular}{c|c|c|c|c|c|c|c|c|c|c|c|c|c|}
\cline{2-14}
\multicolumn{1}{l|}{}                                   & Category                    & bicycle & car & motor & airplane & bus & train & bird & cat & dog & horse & sheep & cow \\ \hline
\multicolumn{1}{|c|}{\multirow{5}{*}{$\mathbb{F}$}} 
& $1^{st}$ layer &         &     &       &           &     &       &       &     &      &       &       &      \\ \cline{2-14} \multicolumn{1}{|c|}{}
& $2^{nd}$ layer &         &     &       &           &     &       &       &     &      &       &       &      \\ \cline{2-14} \multicolumn{1}{|c|}{}
& $3^{rd}$ layer & \checkmark        & \checkmark     & \checkmark       &          & \checkmark    & \checkmark      &      & \checkmark     &     & \checkmark      & \checkmark      &     
            \\ \cline{2-14} \multicolumn{1}{|c|}{}
& $4^{th}$ layer &         &     &       & \checkmark          &     &       & \checkmark      &     & \checkmark     &       &       & \checkmark    
            \\ \cline{2-14} \multicolumn{1}{|c|}{}
& $5^{th}$ layer &         &     &       &           &     &       &       &     &      &       &       &      \\ \hline

\end{tabular}
\end{center}
\caption{\texttt{CFD}'s result on various instances.}
\label{tab:result_cfd}
\end{table*}

\begin{figure*}[hb!]
\begin{center}
\includegraphics[scale=1.0]{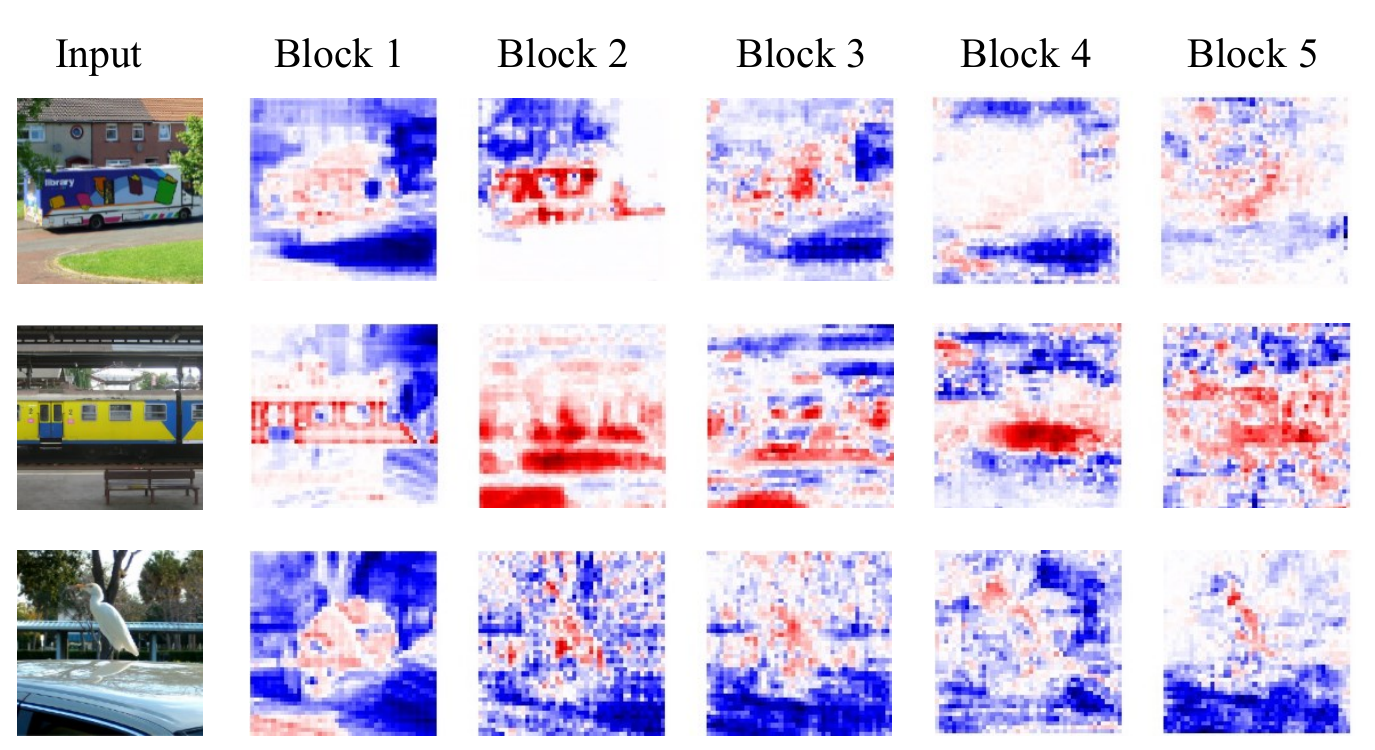}
\end{center}
   \caption{Feature maps from convolutional blocks of ResNet-50.}
\label{fig:Visualizing}
\end{figure*}

\begin{figure*}[bt]
\begin{tabular}{cc}
\bmvaHangBox{\includegraphics[width=7.5cm, height=5.5cm]{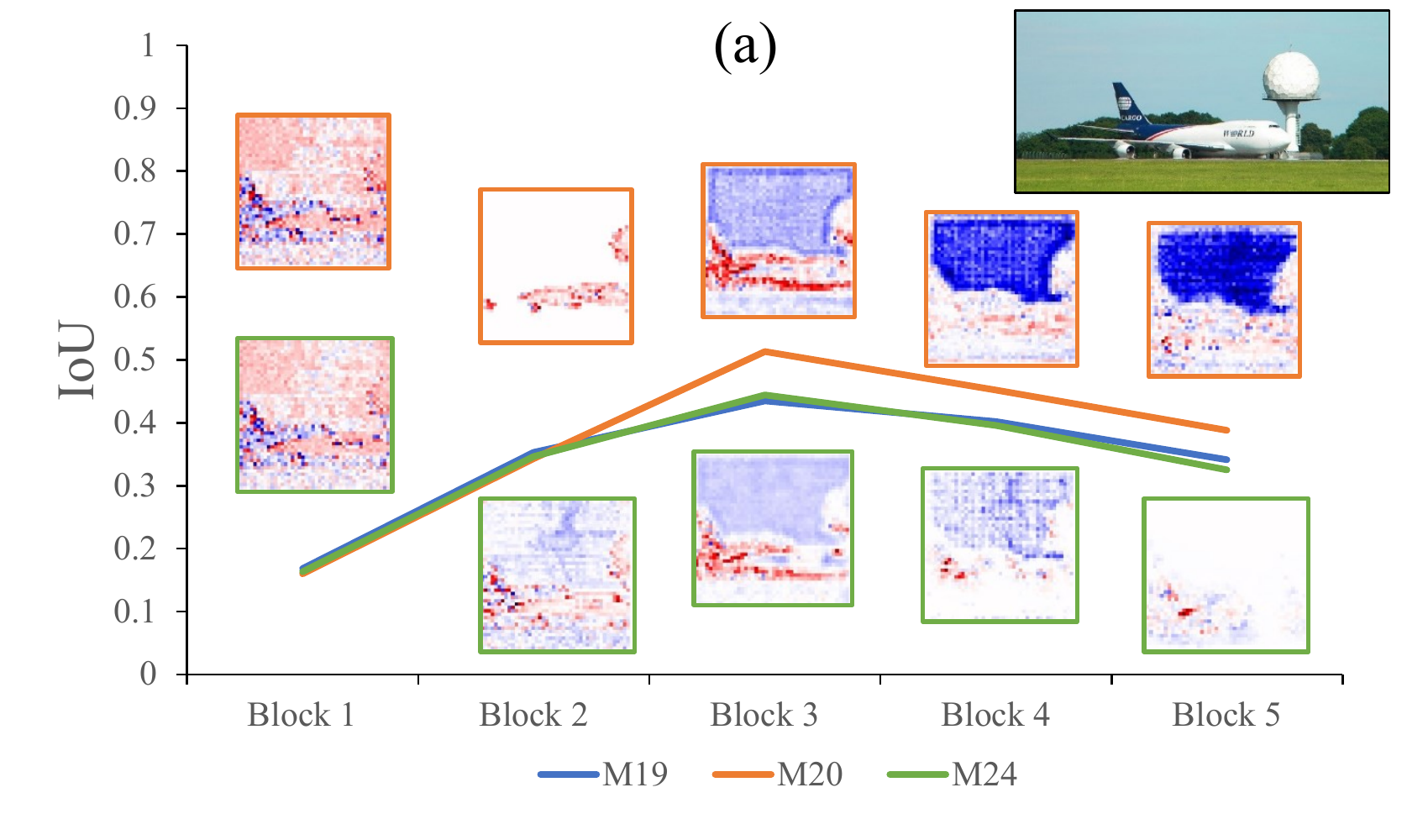}}&
\bmvaHangBox{\includegraphics[width=7.5cm, height=5.5cm]{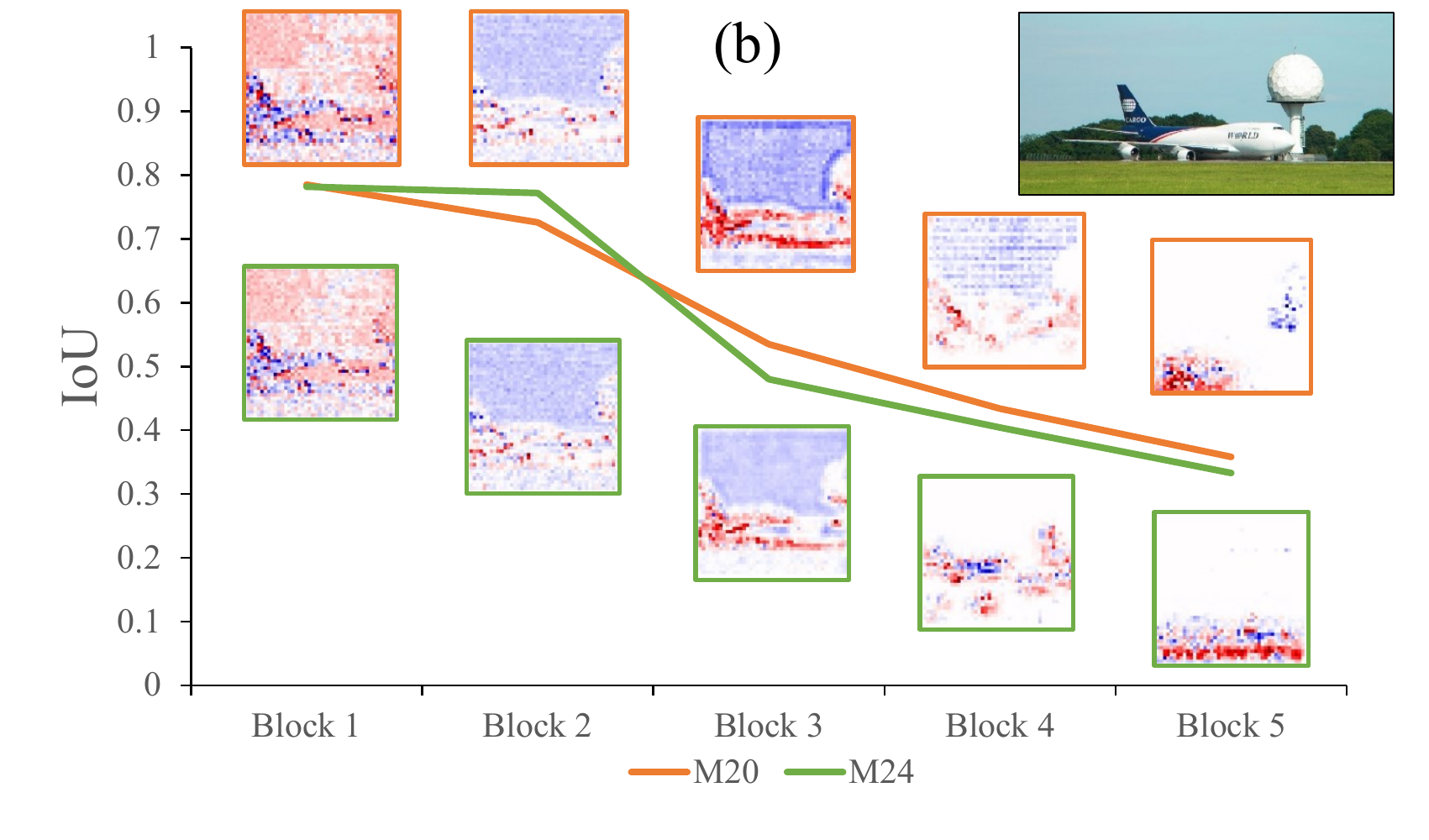}}\\
\end{tabular}
   \caption{(a) $IoU_{M,GT}$ of $M_{19}$, $M_{20}$ and $M_{24}$ comparing with GT. (b) $IoU_{MT,MO}$ of model $M_{20}$ and $M_{24}$ comparing with $M_{19}$.}
\label{fig:IoU_GT}
\end{figure*}

After adding a new class, we obtain $M_{20}$, and $M_{n}$ is the model when a total number of $n$ classes are witnessed. In Fig. \ref{fig:Visualizing}, the visualized results show that the first and second blocks of ResNet-50 can overall capture the outline of objects. Computer vision turns to represent more detailed features from the objects and other background features to determine the class of the input image in the deeper blocks. Also, the IoUs of different blocks in models, comparing with ground truth, are calculated by equation \eqref{eq:1} and equation \eqref{eq:2}. The results reveal that although different models show the performance of the classification inconsistently, the $IoU_{M_n,GT}$ ($n>19$) values are roughly similar at all the layers, which implies that no matter the how good the performance is, the level of matching between feature maps of each model and the human vision is preserved (Fig. \ref{fig:IoU_GT} (a)).

To measure how much the forgetting occurs in the ResNet, we compute IoUs by equation \eqref{eq:3} and equation \eqref{eq:4}. It is clearly shown in Fig. \ref{fig:IoU_GT} (b) that the IoUs between $M_{20}$ and $M_{19}$ are roughly equal to the corresponding figure for $M_{24}$ and $M_{19}$ in every block. For $M_{20}$, the $IoU_{M_{20},M_{19}}$ is always around 0.8 at the first block, showing that a marginal forgetting happens here. The $IoU_{M_{20},M_{19}}$ starts to drop along the blocks because the later feature maps are constructed by the previous maps. The forgetting effect persists and does not show which block is forgetting the most. 

For $M_{24}$, the first block of the model still gets a high IoU comparing with $M_{19}$ and the values decrease from the second block.
Unlike the constantly decreasing trend seen in $M_{20}$, the decreasing rate of $IoU_{M_{24}, M_{19}}$ fluctuates through the blocks and a severe drop at block 3 is observed in every testing input, suggesting that the forgetting effect might happen the most in this block. Iterating this procedure on all images of the set $S$ reinforces that the worst forgetting happens at block 3.

\begin{table*}[hbt]
\begin{center}
\begin{adjustbox}{width=1.0\textwidth,center=\textwidth}
\begin{tabular}{|l|c|c|c|c|c|c|c|c|}
\hline
\multirow{2}{*}{} & \multicolumn{4}{c|}{$Past-task$}                         & 
\multicolumn{4}{c|}{$New-task$}              \\ \cline{2-9} 
                           & BLEU4         & ROUGE\_L       & CIDEr     & SPICE                         & BLEU4        & ROUGE\_L       & CIDEr     & SPICE     \\ \hline
\textbf{Fine-tuning}& 4.2                   & 32.3          & 4.9   & 1.9               & 10.1                    & 39.6         & 17.9      & 5.9    \\ \hline
\textbf{Encoder-Freezing}& 3.8                  & 31.5          & 5.2    & 2.0            & 8.8                    & 38.4          & 15.1      & 5.1    \\ \hline
\textbf{Decoder-Freezing}& 5.2                   & 33.3          & 6.8   & 2.2                                      & 10.6         & 39.8          & 18.5   & 5.5   \\ \hline
\textbf{L1-Freezing}& 5.2       & 33.5         & 7.9    & 2.4         & 12.0      & 41.1          & 23.4      & 6.4    \\ \hline
\textbf{L2-Freezing}& 5.8                   & \textbf{33.6}         & 7.4    & 2.1            & 12.0                   & \textbf{41.2}          & 24.0      & 6.5    \\ \hline
\textbf{L3-Freezing}& 4.5                   & 32.2         & 6.8    & 2.1          & 10.4                   & 39.6          & 20.7      & 6.0    \\ \hline
\textbf{L4-Freezing}& 5.0                   & 32.6         & 7.6    & 2.2            & 11.4                   & 40.0          & 23.3      & 6.2    \\ \hline
\textbf{L5-Freezing}& 5.4                   & 33.0         & 7.4    & 2.3            & 12.1                   & 40.4          & 23.3      & 6.2    \\ \hline
\textbf{LwF\cite{li2017learning}}& \textbf{6.4}                  & 33.2          & \textbf{9.7}   & 2.6                    & 10.6        & 38.9          & 16.1     & 5.3     \\ \hline
\textbf{KD1 \cite{michieli2019incremental}}& 4.7                  & 33.3          & 6.7   & 2.1                    & 11.4        & 40.4          & 20.8     & 6.2     \\ \hline
\textbf{KD2 \cite{michieli2019incremental}}& 3.9                  & 32.4     & 5.6      & 2.0     & 10.6   & 39.8          & 18.6     & 5.9     \\ \hline
\textbf{Critical Freezing}& 5.6                  & 33.2               & 9.5   & \textbf{2.8}                   & \textbf{12.2}                    & 40.7          & \textbf{26.1}       & \textbf{6.9}   \\ \hline
\end{tabular}
\end{adjustbox}
\end{center}
\caption{Performance when 5 classes arrive sequentially on past tasks and newly added tasks.}
\label{tab:my-table}
\end{table*}

While two naive approaches of freezing are also implemented, we devise critical freezing based on findings, which only freezes critical layers. As shown in Table. \ref{tab:my-table}, precisely freezing helps to learn on both the new and old tasks much more effectively. Our freezing scheme outperforms the other approaches on new tasks by a large margin (26.1 CIDEr) while achieving comparable performance with LwF \cite{li2017learning} on past tasks (9.5 CIDEr) although LwF \cite{li2017learning} is far more complicated. Knowledge distillation on intermediate feature space (KD1) and output layer (KD2) \cite{michieli2019incremental} claims 20.8 and 18.6 CIDEr respectively. We argue that the frozen layers contain global information derived from past tasks, which is valuable and should be accumulated during lifetime rather than changing. Fine-tuning optimizes the loss on the new task without any guidance; as a result, the model may not fall into the low-error regions of tasks. We try to freeze each layer of the ResNet to fortify the hypothesis that properly freezing is really better than ambiguous freezing in fine-tuning schemes. Hence, critical freezing exerts a promising influence on fine-tuning techniques in deep learning.     

\begin{figure*}[h!]
\begin{center}
\includegraphics[scale=0.58]{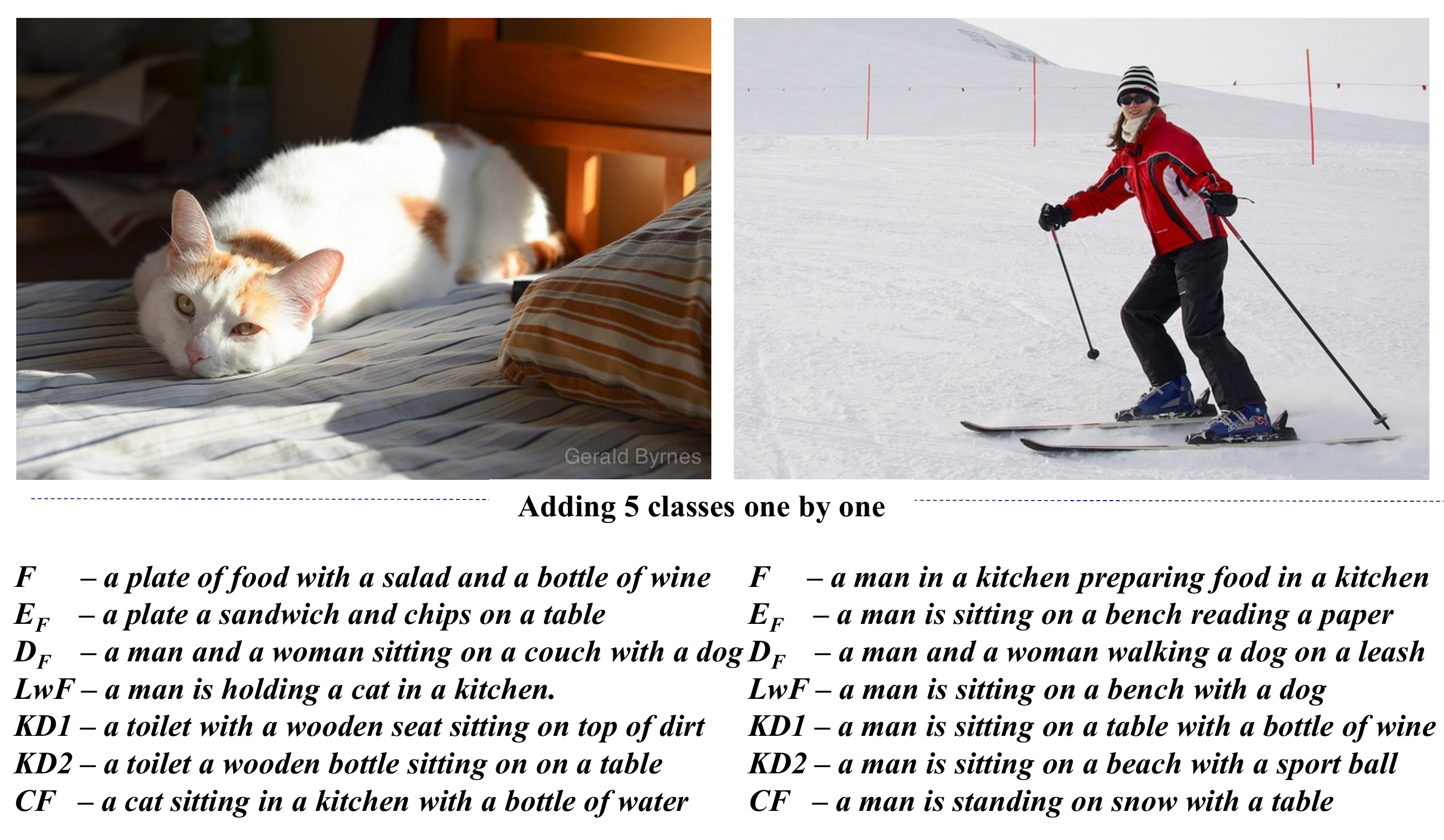}
\end{center}
   \caption{Qualitative results of continual learning algorithms.}
\label{fig:qualitative}
\end{figure*}

Captions are inferred in every setting to qualitatively evaluate the framework (See Fig. \ref{fig:qualitative}). $F$, $E_F$, $D_F$, and $CF$ are fine-tuning, encoder freezing, decoder freezing, and critical freezing respectively. The model misclassifies a cat without using pseudo-labeling or critical freezing when adding 5 classes sequentially. On an image of new classes, critical freezing produces the most relevant caption to the content.  

\newpage
\section{Conclusion and Future Work}
In this work, we introduce a scalable framework, fusing continual learning and image captioning, working on a new dataset called Split MS-COCO created from the standard MS-COCO. We firstly perform image captioning in incremental schemes and then add techniques from continual learning to weaken catastrophic forgetting. Working on the most challenging scenario (class incremental) ensures our framework can perform well in practical schemes while old classes and new classes can co-occur. The experiments in three settings indicate the sign of catastrophic forgetting and the effectiveness when integrating freezing, pseudo-labeling, and distillation.

As the presence of catastrophic forgetting hinders the life-long learning, understanding how this phenomenon happens in computer vision is imperative. We introduce \texttt{CFD} to grasp catastrophic forgetting. The investigation of our tool unearths the mystical question about catastrophic forgetting. 

From knowing where the forgetting issue is coming from, a new technique has been proposed focusing on plastic components of a model to moderate the information loss. The experiments illustrate the superiority of critical freezing over various freezing schemes and existing techniques. To the best of our knowledge, no work has been done for mitigating catastrophic forgetting under the supervision of Interpretable ML. By knowing which regions are needed to be kept intact, not only could the performance on the old task be largely improved, but the new task is also more conquerable. Ultimately, critical freezing could benefit a variety of fine-tuning schemes and continual learning approaches.

There are future works following our research. Firstly, scaling this work for other tasks and deep networks can better validate the feasibility of the proposed continual learning algorithm. Secondly, RNN is now being overlooked and not understood fully via interpretability methods. RNN dissection is arduous but would greatly impact diverse domains (captioning, machine translation, or text synthesis).


\newpage
{\small
\bibliographystyle{unsrt}
\bibliography{references.bib}
}

\end{document}